\documentclass{osa-article}
\pdfoutput=1
\journal{osac}

\articletype{Research Article}

\usepackage{lineno}

\usepackage{cite}
\usepackage{algorithmic}
\usepackage{graphicx}

\usepackage{textcomp}
\usepackage{caption}
\usepackage{subcaption}

\usepackage{float}
\usepackage{booktabs}
\usepackage{multirow}

\usepackage{lipsum,mwe,cuted}

\begin{document}

\title{Light field Rectification based on relative pose estimation}

\author{Xiao Huo,\authormark{1} Dongyang Jin,\authormark{2}, Saiping Zhang,\authormark{1} and Fuzheng Yang\authormark{1,*}}

\address{\authormark{1}State Key Laboratory of Integrated Services Networks, Xidian University, Xi’an, 710071, China\\
\authormark{2}College of Information and Communication, National University of Defense Technology Xi’an 710106, China }

\email{\authormark{*}fzhyang@mail.xidian.edu.cn} 



\begin{abstract}
Hand-held light field (LF) cameras have unique advantages in computer vision such as 3D scene reconstruction and depth estimation. However, the related applications are limited by the ultra-small baseline, e.g., leading to the extremely low depth resolution in reconstruction. To solve this problem, we propose to rectify LF to obtain a large baseline. Specifically, the proposed method aligns two LFs captured by two hand-held LF cameras with a random relative pose, and extracts the corresponding row-aligned sub-aperture images (SAIs) to obtain an LF with a large baseline. For an accurate rectification, a method for pose estimation is also proposed, where the relative rotation and translation between the two LF cameras are estimated. The proposed pose estimation minimizes the degree of freedom (DoF) in the LF-point-LF-point correspondence model and explicitly solves this model in a linear way. The proposed pose estimation outperforms the state-of-the-art algorithms by providing more accurate results to support rectification. The significantly improved depth resolution in 3D reconstruction demonstrates the effectiveness of the proposed LF rectification.
\end{abstract}

\section{Introduction}
The commercialization of light field (LF) cameras, such as Lytro~\cite{Lytroorg} and Raytrix~\cite{Raytrixorg}, have accelerated the development of LF technologies. Unlike the classic pinhole camera, an LF camera can capture both spatial and angular information of rays in the environment by a single 2D snapshot, which is the reason why the LF camera has gained significant interest in computer vision. Typical applications of the LF camera include simultaneous localization and mapping (SLAM)~\cite{dong2013plenoptic,zeller2017calibration}, 3D reconstruction~\cite{johannsen2016layered,Wu_2017_CVPR,Nousias_2019_CVPR}, digital refocusing ~\cite{ng2005fourier}, virtual viewpoint synthesis~\cite{10.1145/2980179.2980251}, super resolution~\cite{Jin_2020_CVPR,meng2020high}, etc.

\begin{figure}[htb]
	\centering
	\includegraphics[width=60mm]
	{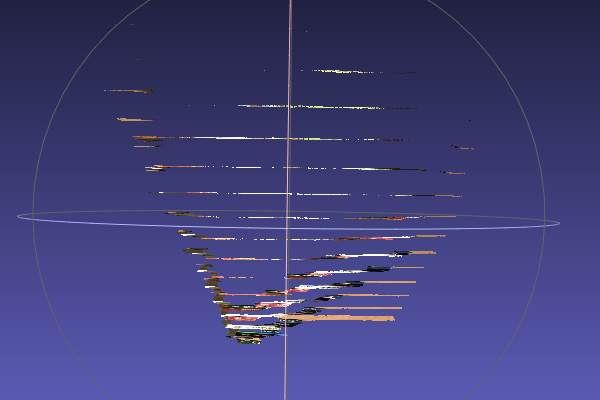}
	\caption{The top view of a layered 3D reconstruction from a single LF.}
	\label{fig:intro}
 \end{figure}

One of the most important tasks of the LF camera is depth estimation~\cite{jeon2015accurate,tao2013depth,wanner2012globally,8985549,Huang_2017_ICCV,rogge2020depth,8263242}. Compared with traditional stereo image pairs, LFs extend disparity to a continuous space. This advantage is apparent when considering epipolar plane images (EPIs)~\cite{bolles1987epipolarp}. Due to a dense sampling in the angular direction (e.g., 15 15 sub-views in Lytro Illum), corresponding pixels of a scene point in sub-aperture images (SAIs) can be projected onto a slope line in EPIs, and line parameters can be encoded into dense stereo matching to obtain a more robust depth estimation~\cite{wanner2012globally}. However, a single LF has a weakness which cannot be ignored: the extremely small baseline between SAIs (e.g., 14  in Lytro Illum). With this structural imperfection, even if LF depth estimation methods~\cite{jeon2015accurate,8985549,Huang_2017_ICCV,rogge2020depth,8263242} achieve a high disparity resolution, a low depth resolution is still inevitable, which generally leads to a layered 3D reconstruction, as shown in Fig.~\ref{fig:intro}. To obtain an LF with a larger baseline, on one hand, using the camera array~\cite{wilburn2005high,1315176} is a candidate choice. However, it is complicated to align all cameras in the camera array and high in cost to reach the identical angular resolution as that of an LF camera. On the other hand, manually aligning two hand-held LF cameras is another straightforward solution, whereas it is also hard to make the image planes reside in the same plane and the rows of corresponding SAIs completely aligned. Hence it is critical to mathematically align two hand-held LF cameras with random relative pose to obtain an LF with a large baseline.

In order to achieve an accurate alignment, the relative pose is first estimated. Relative pose estimation has attracted a lot attention in research. The most general framework~\cite{1211520} defined a generalized camera model as a collection of rays, and established the correspondences between rays intersecting the same 3D point, denoted as the generalized epipolar constraint (GEC). Taking all kinds of degenerated camera configurations into consideration, Li et al.~\cite{4587545} used a linear algorithm to solve the relative pose from the GEC efficiently without ambiguities, which was also applicable to LF cameras. Johannsen et al.~\cite{johannsen2015on} first introduced the GEC into the LF camera and considered the geometric constraints between projections within a single LF, which effectively utilized the characteristics of the LF camera.

However, the rays reconstructed by the small baseline are very dense and easy to be corrupted by noises in the above mentioned methods~\cite{9320357}. Accordingly, Nousias et al.~\cite{9320357} proposed an LF projection matrix to represent the correspondence between LF features and 3D points, and used the direct linear transformation (DLT) to estimate the absolute pose of LF cameras. In our previous work~\cite{9694504}, the LF-point-LF-point correspondence (LLC) model was established, which described the correspondence between corresponding LF features (LF-points) from a pair of LFs. Rotation and translation were directly decoupled from the model and solved separately. In this paper, we explicitly solved the parameters in the LLC model, from which more accurate of relative pose is extracted. Based on the extracted relative pose, we establish a pairwise rays warping function, and construct the rectification matrixes and vectors to align LFs captured by two hand-held LF cameras so that an LF with a large baseline can be obtained. A subsequent quadrilinear interpolation is implemented to acquire the row-aligned SAI sets. By rectification, the dimension of searching space in depth estimation is reduced from two to one, which directly makes stereo matching in different LFs feasible. A commonly used depth estimation algorithm designed for a single LF captured by an LF camera is performed on the rectified LF to output the depth in a high depth resolution.
 
The main contributions include:

(1) We propose a linear approach without redundant variables to solve the LF-point-LF-point correspondence and recover the relative pose from it. Experimental results demonstrate that, compared with other state-of-the-art relative pose estimation algorithms, our estimated relative poses are both accurate and robust, which prepares for the proposed LF rectification.

(2) To the best of our knowledge, our work is the first attempt to rectify two 4D LFs based on the estimated relative pose of LF cameras to obtain an LF with row-aligned SAIs and a larger baseline.

(3) We build a rectification pipeline on LFs captured by two LF cameras which can support to generate 3D reconstructions in high depth resolution.

\section{LF rectification pipeline}
\begin{figure}[htbp]
	\centering
	\includegraphics[width=130mm]
	{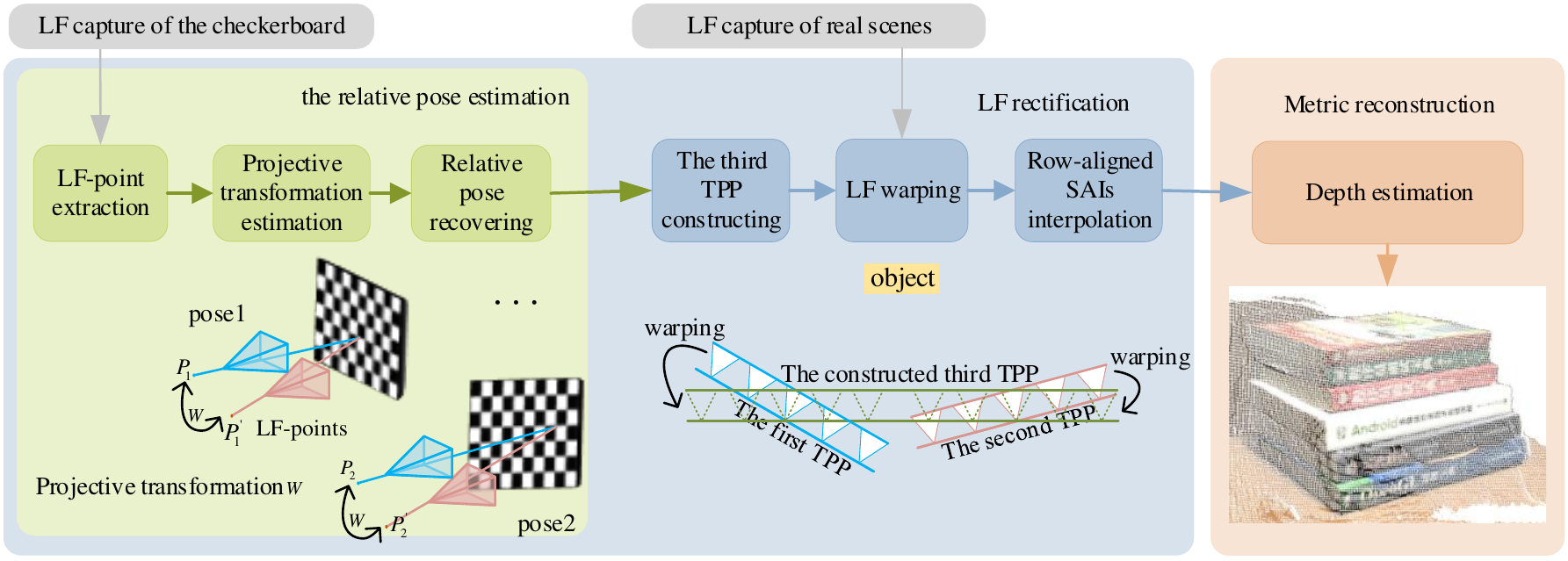}
	\caption{The pipeline of LF rectification.}
	\label{fig:flow}
\end{figure}
Fig.~\ref{fig:flow} shows the pipeline of the proposed LF rectification. The LFs of the checkerboard and real scenes are captured by two internal-calibrated LF cameras with fixed relative pose. As in our previous work~\cite{9694504}, the calibration is performed using the method introduced in~\cite{2020arXiv200103734J}. To avoid the degeneracy in the LLC model, the checkerboard should be placed in different poses. Details will be introduced in Sec.~\ref{solution}.
 
In the relative pose estimation, the LF-points of different corners in two cameras are extracted respectively by the approach introduced in~\cite{2020arXiv200103734J}. Then these LF-points pairs are normalized based on the invariance to image coordinate transformation. The parameters in the LLC model are estimated by the proposed linear approach using these LF-point pairs. Finally, the initial relative pose is extracted from these parameters and refined by the Levenberg-Marquardt-on-Manifold algorithm. Details of this part will be introduced in Sec.~\ref{poseEsti}.

In the LF rectification, the LF pairs of real scenes are processed. In the preparation phase, for a pair of LFs, SAI sets of both LFs are acquired respectively through the extraction algorithm in~\cite{bok2016geometric}. A LF camera can be regarded as a set of pinhole cameras arranged in a gird~\cite{bok2016geometric} (whose SAIs after calibration are equivalent to those images captured by these pinhole cameras) and the simplified 2D case is depicted in Fig.~\ref{fig:flow}. Coordinates and luminance of rays are converted from SAIs through calibrated intrinsic parameters. After obtaining the solved relative pose, rectification starts from constructing the rectification matrices and vectors of the two LFs. Then, based on the warping function represented by the elements of the rectification matrix and vector, the rays of two LFs are warped onto the constructed two-plane parameterized (TPP)~\cite{levoy1996light,gortler1996lumigraph}. Finally, considering that the LF under TPP is a 4D signal, the row-aligned SAIs are obtained through quadrilinear interpolation. These extended SAIs are applied in the subsequent depth estimation. Details will be introduced in Sec.~\ref{recti}.

In the metric reconstruction, the depth estimation is implemented under the framework~\cite{jeon2015accurate}, which estimates the depths through stereo matching within a single LF. Based on the phase shift theorem, SAIs can be displaced with sub-pixel accuracy to calculate the cost volume. After rectification, the depth estimation is performed between the row-aligned SAI sets. With depth of the scene, the point cloud can be reconstructed using the central SAI and intrinsic parameters of the LF camera. Final metric reconstructions exhibited in experiments show that the depth resolution is significantly improved.

\section{Relative pose estimation}\label{poseEsti}
For an effective LF rectification, the prerequisite is to obtain the accurate relative pose of the two LF cameras. If the point or rays outside the LF camera are used to solve the pose, the feature points in the LF images must be used to reconstruct these points or rays through the intrinsic parameters of the LF camera. However, the small baseline, which belongs to intrinsic parameters~\cite{jeon2015accurate}, significantly penalizes the reconstruction accuracy of rays and points. For rays, the sampling interval is usually only one hundredth of a pixel, which makes it difficult to distinguish them, resulting in significant errors in reconstruction accuracy. More analysis can be found in~\cite{9320357}. For 3D points, the small baseline will amplify the detection noise of features (e.g., 2D SIFT feature). The depth $Z$ of points can be calculated by depth formula as
\begin{equation}
   Z=\frac{B f}{D},
\end{equation}
where $f$ is the focal length of the LF camera, and $D$ is the disparity of features between the adjacent SAIs. The differential change of $Z$ is calculated as
\begin{equation}\label{dZ}
   d Z=-\frac{Z^{2}}{B f} d D.
\end{equation}

For a baseline $B=14\mu m$,  a small disparity error $dD$ can make for a large depth error $dZ$. Here the disparity is expressed by the coordinate difference of detected feature points in two adjacent SAIs. Therefore, the reconstructed rays and points do not provide valuable information~\cite{9320357}.

\subsection{Preliminary}
\subsubsection{LF-point}\label{LFP}
LF-point $\left(u_{c}, v_{c}, \lambda\right)$ is an LF feature first defined in~\cite{2020arXiv200103734J} which has one-to-one correlation with the 3D scene point. The mathematical expression is composed of the projection coordinates $\left(u_{c}, v_{c}\right)$ of the 3D point in the central SAI and its disparity $\lambda$ between any two adjacent SAIs. According to~\cite{bok2016geometric}, the sub-aperture images are evenly arranged in the image plane, thus the disparities of the 3D point obtained from different adjacent SAIs are assumed to be equal. It indicates that this triplet can completely contain the inherent information about the LF structure of 3D scene points.
\subsubsection{LF-point-LF-point correspondence model}
As defined in our previous work~\cite{9694504}, the LF-point-LF-point correspondence model is established as
\begin{equation}
\label{LFPM}
\left[\begin{matrix}
u_{c}^{\prime} \\
v_{c}^{\prime} \\
\lambda^{\prime} \\
1
\end{matrix}\right]=\underbrace{\frac{Z}{Z^{\prime}} \underbrace{\underbrace{\left[\begin{matrix}
f_{x}^{\prime} & 0 & c_{x}^{\prime} & 0 \\
0 & f_{y}^{\prime} & c_{y}^{\prime} & 0 \\
0 & 0 & -K_{1}^{\prime} & -K_{2}^{\prime} \\
0 & 0 & 1 & 0
\end{matrix}\right]}_{H^\prime}\left[\begin{matrix}
R_{3 \times 3} & T_{3 \times 1} \\
0_{1 \times 3} & 1
\end{matrix}\right] \underbrace{\left[\begin{matrix}
\frac{1}{f_{x}} & 0 & 0 & -\frac{c_{x}}{f_{x}} \\
0 & \frac{1}{f_{y}} & 0 & -\frac{c_{y}}{f_{y}} \\
0 & 0 & 0 & 1 \\
0 & 0 & -\frac{1}{K_{2}} & -\frac{K_1}{K_{2}}
\end{matrix}\right]}_{H^{-1}}}_{W}}_{M}\left[\begin{matrix}
u_{c} \\
v_{c} \\
\lambda \\
1
\end{matrix}\right],
\end{equation}
where ${\left[ {{u_c},{v_c},\lambda ,1} \right]^\top}$ and ${\left[ {u_c^\prime ,v_c^\prime ,\lambda^\prime ,1} \right]^\top}$ are the homogeneous coordinates of LF-points (in a pair of LFs) corresponding to the same 3D point. $Z$ and $Z^\prime$ are the depth of the 3D point in the two LF camera coordinate systems, respectively. $H$ and $H^\prime$ are the intrinsic parameter matrices of the two LF cameras which are already known after calibration. ${R_{3 \times 3}}$ and ${T_{3 \times 1}}$ are the unknown extrinsic parameters which need to be estimated. This correspondence indicates that using LF-point pairs to solve the relative pose, the difficulties and loss of accuracy in reconstructing rays or 3D points can be avoided. 

\subsection{Proposed solution of the relative pose}\label{solution}
Eq.~\eqref{LFPM} shows that the matrix $M$ can map a LF-point to its corresponding LF-point. For different scene points, their LF-points can be mapped to the corresponding LF-points by the same projective transformation $W$ under homogeneous coordinates, since $W$ only relates to the intrinsic and extrinsic parameters of these two cameras. 

Based on the above analysis, the projective transformation can be expressed in the matrix form as
\begin{equation}\label{proj}
	P^{\prime} \simeq W P,
\end{equation}
where $P$ and $P^{\prime}$ are a pair of LF-points, and $\simeq$ denotes equality up to a scale factor. 

In data-preprocessing, the LF-points are normalized based on the invariance to image coordinate transformation~\cite{hartley2003multiple}. The normalizing matrix $N$ is introduced, and the LF-point becomes
\begin{equation}\label{norm}
	\bar{P}=\underbrace{\left[\begin{matrix}v_{1} & 0 & 0 & x_{1} \\ 0 & v_{2} & 0 & x_{2} \\ 0 & 0 & v_{3} & x_{3} \\ 0 & 0 & 0 & 1\end{matrix}\right]}_{N}P,
\end{equation}
where $v_1$, $v_2$, and $v_3$ are scaling factors for image and disparity coordinate. $\left[x_{1}, x_{2}, x_{3}\right]^\top$ are the translation vector. Normalization translates the set of LF-points so as to bring the centroid of them to the origin and scale them to make the two principal moments both equal to unity.  An independent normalization is also applied to another set of LF-points, and $N^{\prime}$ has a similar form.

Substitute Eq.~\eqref{norm} into Eq.~\eqref{proj}, we obtain
\begin{equation}\label{normed_proj}
	\bar{P}^{\prime}=\underbrace{cN^{\prime} W N^{-1} }_{W^\prime}\bar{P},
\end{equation}
where $c$ is an unknown scalar. Based on the DLT algorithm described in~\cite{hartley2003multiple},  Eq.~\eqref{normed_proj} can be expanded by taking the cross product of the left and right-hand sides as:

\begin{equation}\label{g_n_s_proj}
	A\operatorname{vec}\left(W_{16}^{\prime}\right)=0,
\end{equation}
where the matrix elements of $A$ are quadratic in the known coordinates of the LF-points, and $\operatorname{vec}\left(W_{16}^{\prime}\right)$ is a $16 \times 1$ vector consisting of the entries of the matrix $W^{\prime}$.

$W^{\prime}$ can be solved from Eq.~\eqref{g_n_s_proj} by performing singular value decomposition (SVD), and then the rotation and translation can be extracted from $W^{\prime}$. On the other hand, $W^{\prime}$ represents a 3D projective transformation which can only be defined up to scale. The total number of DoF of it is 15. The Rotation (considered as 9 independent variables) and translation take 12 DoF, thus there are still redundant variables in Eq.~\eqref{g_n_s_proj}. Here, we give 3 linear constraints to minimize the DoF as
\begin{equation}
 \begin{cases}
w_{9}^{\prime}=\alpha^{\prime} w_{13}^{\prime}\\
	w_{10}^{\prime}=\alpha^{\prime} w_{14}^{\prime}\\
	w_{12}^{\prime}=\alpha^{\prime} w_{16}^{\prime}+\alpha w_{0}^{\prime}
\end{cases},   
\end{equation}
where $w_{0}^{\prime}=-w_{11}^{\prime}+\alpha^{\prime} w_{15}^{\prime}$, $\alpha^{\prime}=x_{3}^{\prime}-K_{1}^{\prime} v_{3}^{\prime}$ and $\alpha=x_{3}-K_{1} v_{3}$. The derivation can be found in section 1 in Supplement 1. Then $\operatorname{vec}\left(W_{16}^{\prime}\right)$ can be represented as
 \begin{equation}\label{w16_w13}
	\left[\begin{matrix}
	w_{1}^{\prime} \\
	w_{2}^{\prime} \\
	\vdots \\
	w_{16}
	\end{matrix}\right]=\underbrace{\left[\begin{matrix}
	I_{8 \times 8} &  & &0_{8 \times 5} \\
	0_{1 \times 8} & \alpha & 0 & 0 & 0 & 0 \\
	0_{1 \times 8} & 0 & \alpha^{\prime} & 0 & 0 & 0 \\
	0_{1 \times 8} & 0 & 0 & \alpha^{\prime} & 0 & -1 \\
	0_{1 \times 8} & 0 & 0 & 0 & \alpha^{\prime} & \alpha \\
	0_{4 \times 8} & & & I_{4 \times 4}& & 0_{4 \times 1}
	\end{matrix}\right]}_{Q}\left[\begin{matrix}
	w_{1}^{\prime} \\
	w_{2}^{\prime} \\
	\vdots \\
	w_{8}^{\prime} \\
	w_{13}^{\prime} \\
	\vdots \\
	w_{16}^{\prime} \\
	w_{0}^{\prime}
	\end{matrix}\right].
	\end{equation}
	
The coefficient matrix in Eq.~\eqref{w16_w13} is represented by $Q$, and variables on the right side of the equation are represented by $\operatorname{vec}\left(W_{13}^{\prime}\right)$. $I$ is the identity matrix, and the subscript indicates its size. Substituting~\eqref{w16_w13} to~\eqref{g_n_s_proj}, we obtain
 \begin{equation}\label{g_n_s_proj_13}
	A Q \operatorname{vec}\left(W_{13}^{\prime}\right)=0.
 \end{equation}

By performing SVD on $AQ$, $\operatorname{vec}\left(W_{13}^{\prime}\right)$ is solved subject to $\Vert\operatorname{vec}\left(W_{13}^{\prime}\right) = 1\Vert$. Then we calculate $\operatorname{vec}\left( W_{16}^{\prime} \right)$, reform it to matrix $W^\prime$ up to a scalar $\mu$, de-normalize $W^\prime$, and remove the intrinsic parameter matrices to calculate the rotation and translation. This process is expressed through rearranging $W^{\prime}$ in Eq.~\eqref{normed_proj} as,

\begin{equation}
	\frac{1}{\mu}\left[\begin{matrix}
	R_{3 \times 3} & T_{3 \times 1} \\
	0_{1 \times 3} & 1
	\end{matrix}\right]=H^{\prime-1} N^{\prime-1} W^{\prime} N H.
\end{equation}

$R_{3 \times 3}$ and $T_{3 \times 1}$ can be obtained through scaling the matrix on the left-hand side.

Considering the orthogonality of rotation matrix, $R$ is projected into the space of rotation matrices (SO(3)) using the method in~\cite{Horn:87} to get its initial solution. 

Referring~\cite{9694504}, we restate Eq.~\eqref{LFPM} to a linear form as
\begin{equation}\label{solveT}
	\left[\begin{matrix}
	A_{R} & A_{T}
	\end{matrix}\right]\left[\begin{matrix}
	\operatorname{vec}\left(R_{3 \times 3}\right) \\
	\operatorname{vec}\left(T_{3 \times 1}\right)
	\end{matrix}\right]=0,
 \end{equation}
where $A_{R}$ and $A_{T}$ are the known coefficient matrix of $\operatorname{vec}\left(R_{3 \times 3}\right)$ and $\operatorname{vec}\left(T_{3 \times 1}\right)$ respectively. $\operatorname{vec}\left(R_{3 \times 3}\right)$ is the $9 \times 1$ vector formed by stacking the columns of $R$, and $\operatorname{vec}\left(T_{3 \times 1}\right)$ is exactly $T$. We substitute $R$ into Eq.~\eqref{solveT} to obtain the initial solution of translation vector which is calculated as,
 \begin{equation}
	\operatorname{vec}\left(T_{3 \times 1}\right)=-A_{T}^{+} A_{R} \operatorname{vec}\left(R_{3 \times 3}\right),
 \end{equation}
where $A_{T}^{+}$ represents the pseudo-inverse of  $A_{T}$.

It should be noted that when all 3D points are coplanar, there may be degeneracy in the LLC model. This situation is given as 
\begin{equation}\label{degeneracy}
	\left[\begin{matrix}
	u_{c}^{\prime} \\
	v_{c}^{\prime} \\
	\lambda^{\prime} \\
	1
	\end{matrix}\right]=\frac{Z}{Z^{\prime}}\underbrace{ H^{\prime}\left[\begin{matrix}
	R_{3 \times 3}-\frac{T_{3 \times 1} n^\top}{d} & 0_{3 \times 1} \\
	0_{1 \times 3} & 1
	\end{matrix}\right] H^{-1}}_{U}\left[\begin{matrix}
	u_{c} \\
	v_{c} \\
	\lambda \\
	1
	\end{matrix}\right],
 \end{equation}
where $n$ denotes the normal vector of this plane, and $d$ is the distance from plane to the origin. Eq.~\eqref{degeneracy} shows that $U$ is another solution of the homography matrix. The derivation of Eq.~\eqref{degeneracy} can be found in section 2 in Supplement 1.

The initial solution computed by the linear method is refined via non-linear optimization. The variables to be optimized are pose parameters $R$ and $T$. The reprojection cost function measuring the geometric distance between the LF-point is defined as follow:
 \begin{equation}\label{BA}
	\sum_{i=1}^{N} d\left(\left[u^{\prime i}, v^{\prime i}, \lambda^{\prime i}\right]^{\mathrm{T}},\left[u_{e s t}^{\prime i}, v_{e s t}^{i}, \lambda_{e s t}^{\prime i}\right]^{\mathrm{T}}\right)^{2},
\end{equation}
where $N$ is the number of LF-points, and $d(\cdot, \cdot)$ is the Euclidean distance. $\left[u_{c}^{\prime i}, v_{c}^{\prime i}, \lambda^{\prime i}\right]^{\mathrm{T}}$ is the LF-point of the  $i$-th 3D point in the second LF. $\left[u_{est}^{\prime i}, v_{est}^{i} t_{est}^{\prime i}\right]^{\mathrm{T}}$ is the corresponding estimated LF-point calculated by Eq.~\eqref{LFPM}. Eq.~\eqref{BA} is minimized to determine the final pose parameters of the camera setup. The minimization is performed by the Levenberg-Marquardt-on-Manifold Algorithm used in~\cite{9694504,strasdat2010real,engel2014lsd,zeller2018scale}.

\section{LF rectification}\label{recti}

\begin{figure}[htbp]
   \centering
   \includegraphics[width=80mm]
   {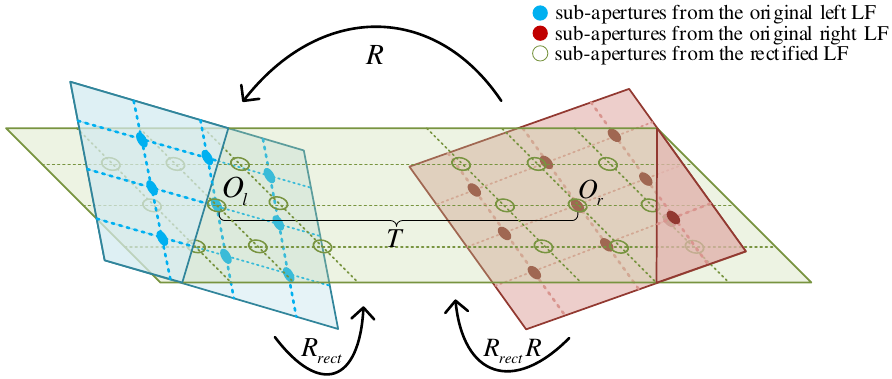}
   \caption{Warping two LFs into TPP parameterization (only main planes where sub-apertures are located shown for better visualization).}
   \label{fig:rec}
\end{figure}

After obtaining the relative pose, one LF can be warped into the TPP parameterization of another LF. However, the direct warping result usually cannot enlarge the baseline effectively. According to the classical work of LF camera calibration~\cite{bok2016geometric}, sub-apertures are located on the main plane of the LF camera. The same row of sub-apertures should have the same vertical coordinates, and the same column of sub-apertures should have the same horizontal coordinates as shown in Fig.~\ref{fig:rec}. Unfortunately, it is hard for users to manually place the two cameras in proper position so that their main planes can be perfectly coplanar and sub-apertures distribute appropriately. The proposed LF rectification allows users to freely rotate and translate the LF cameras and align the two LFs into a common TPP parameterization. Fig.~\ref{fig:rec} shows the goal of LF rectification: constructing the third TPP and warping two LFs into the common TPP parameterization with sub-apertures aligned. 

\subsection{LF warping}\label{warping}
\begin{figure}[htbp]
	\centering
	\includegraphics[width=80mm]
	{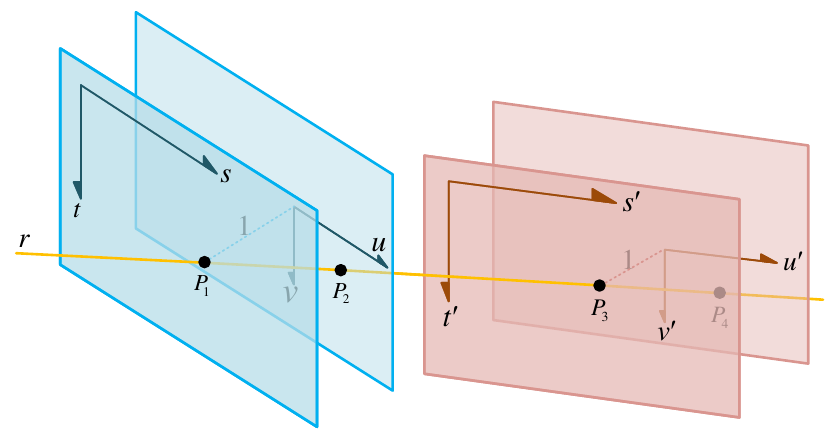}
	\caption{Mapping each ray from $L$ to $L^\prime$.}
	\label{TPP}
 \end{figure}
 
The conventional TPP parameterization which contains the $ST$ plane and the $UV$ plane is used to represent an LF. The intersection point coordinate of the ray and the $ST$ plane is $\left[s,t\right]^\top$, indicating the position of the ray and the sub-aperture to which the ray belongs. The intersection point coordinate of the ray and the $UV$ plane relative to $\left[s,t\right]^\top$ is $\left[u,v\right]^\top$, indicating the direction of the ray, so the ray can be represented by $\left[s,t,u,v\right]^\top$. In this model, we place the $ST$ plane in the main lens plane and the $UV$ plane one unit in front of the $ST$ plane. 
 
To warp the LF, all the rays need to be warped. Given the rotation $R$ and translation $T$, we start with deriving the pairwise rays warping function. Without loss of generality, one light field $L$ is set as the global LF which can be represented by the set of each ray $r$ in it. For each ray, its corresponding coordinate $\left[s^{\prime},t^{\prime},u^{\prime},v^{\prime}\right]^\top$ in the second light field $L^{\prime}$ is computed. This is done by intersecting $r$ with the TPP of $L^{\prime}$ as shown in Fig.~\ref{TPP}. The calculation is followed as
 \begin{equation}\label{ps1}
	\left[\begin{matrix}
	X_{s_{1}} \\
	Y_{s_{1}} \\
	Z_{s_{1}}
	\end{matrix}\right]
	=R \left[\begin{matrix}
	s \\
	t \\
	0
	\end{matrix}\right]+T
 \end{equation}
and 
 \begin{equation}\label{ps2}
	\left[\begin{matrix}
	X_{s_{2}} \\
	Y_{s_{2}} \\
	Z_{s_{2}}
	\end{matrix}\right]
	=R \left[\begin{matrix}
	s+t \\
	v+u \\
	1
	\end{matrix}\right]+T,
 \end{equation}
where $[s, t, 0]^\top$ is the coordinate of $P_{1}$, i.e., the intersection of $r$ and the $ST$ plane in $L$. $[s+u, t+v, 1]^\top$ is the coordinate of $P_{2}$, i.e., the intersection  of $r$ and the $UV$ plane. $\left[X_{s_{1}}, Y_{s_{1}}, Z_{s_{1}}\right]^\top$ and $\left[X_{s_{2}}, Y_{s_{2}}, Z_{s_{2}}\right]^\top$ are used to represent the coordinates of $P_{1}$ and $P_{2}$ in $L^{\prime}$. Then we calculate the coordinates of  $P_{3}$ and $P_{4}$, which are the intersections of $r$ with the TPP of $L^{\prime}$, represented by $\left[s^{\prime}, t^{\prime}, 0\right]^\top$ and $\left[s^{\prime}+u^{\prime}, t^{\prime}+v^{\prime}, 1\right]^\top$, respectively. The coordinates are calculated as
 \begin{equation}\label{ps3}
	\left[\begin{matrix}
	s^\prime \\
	t^\prime \\
	0
	\end{matrix}\right]
	=\left[\begin{matrix}
	X_{s_{1}} \\
	Y_{s_{1}} \\
	Z_{s_{1}}
	\end{matrix}\right]+\lambda_{1}\left(\left[\begin{matrix}
	X_{s_{2}} \\
	Y_{s_{2}} \\
	Z_{s_{2}}
	\end{matrix}\right]-\left[\begin{matrix}
	X_{s_{1}} \\
	Y_{s_{1}} \\
	Z_{s_{1}}
	\end{matrix}\right]\right)
 \end{equation}
 and
 \begin{equation}\label{ps4}
	\left[\begin{matrix}
	s^\prime+u^\prime \\
	t^\prime+v^\prime \\
	1
	\end{matrix}\right]=\left[\begin{matrix}
	X_{s_{1}} \\
	Y_{s_{1}} \\
	Z_{s_{1}}
	\end{matrix}\right]+\lambda_{2}\left(\left[\begin{matrix}
	X_{s_{2}} \\
	Y_{s_{2}} \\
	Z_{s_{2}}
	\end{matrix}\right]-\left[\begin{matrix}
	X_{s_{1}} \\
	Y_{s_{1}} \\
	Z_{s_{1}}
	\end{matrix}\right]\right),
 \end{equation}
 where $\lambda_{1}=\frac{Z_{s_{1}}}{Z_{s_{1}}-Z_{s_{2}}}$ and $\lambda_{2}=\frac{Z_{s_{1}}-1}{Z_{s_{1}}-Z_{s_{2}}}$. Finally, we substitute~\eqref{ps1} and~\eqref{ps2} into~\eqref{ps3} and~\eqref{ps4} and obtain
\begin{equation}
\left[\begin{matrix}
	   {{s}^\prime}  \\
	   {{t}^\prime}  \\
	   {{u}^\prime}  \\
	   {{v}^\prime}  \\
\end{matrix} \right]=\left[ \begin{matrix}
	   {{t}_{1}}+{{r}_{1,1}}s+{{r}_{1,2}}t-\dfrac{\left(t_3+r_{3,1}s+r_{3,2}t\right)\left( {{r}_{1,3}}+{{r}_{1,1}}\left( s+u \right)+{{r}_{1,2}}\left( t+v \right)-{{r}_{1,1}}s-{{r}_{1,2}}t \right)}{{{r}_{3,3}}+{{r}_{3,1}}\left( s+u \right)+{{r}_{3,2}}\left( t+v \right)-{{r}_{3,1}}s-{{r}_{3,2}}t}  \\
	   {{t}_{2}}+{{r}_{2,1}}s+{{r}_{2,2}}t-\dfrac{\left(t_3+r_{3,1}s+r_{3,2}t\right)\left( {{r}_{2,3}}+{{r}_{2,1}}\left( s+u \right)+{{r}_{2,2}}\left( t+v \right)-{{r}_{2,1}}s-{{r}_{2,2}}t \right)}{{{r}_{3,3}}+{{r}_{3,1}}\left( s+u \right)+{{r}_{3,2}}\left( t+v \right)-{{r}_{3,1}}s-{{r}_{3,2}}t}  \\
	   \dfrac{{{r}_{1,3}}+{{r}_{1,1}}u+{{r}_{1,2}}v}{{{r}_{3,3}}+{{r}_{3,1}}u+{{r}_{3,2}}v}  \\
	   \dfrac{{{r}_{2,3}}+{{r}_{2,1}}u+{{r}_{2,2}}v}{{{r}_{3,3}}+{{r}_{3,1}}u+{{r}_{3,2}}v}  \\
	\end{matrix} \right],
\end{equation}
where subscripts indicate the components of the rotation matrix or translation vector.
 
\begin{figure}[htbp]
	\centering
	\subcaptionbox{\label{fig:warping_a}}
	{
	\includegraphics[width=60mm]
	{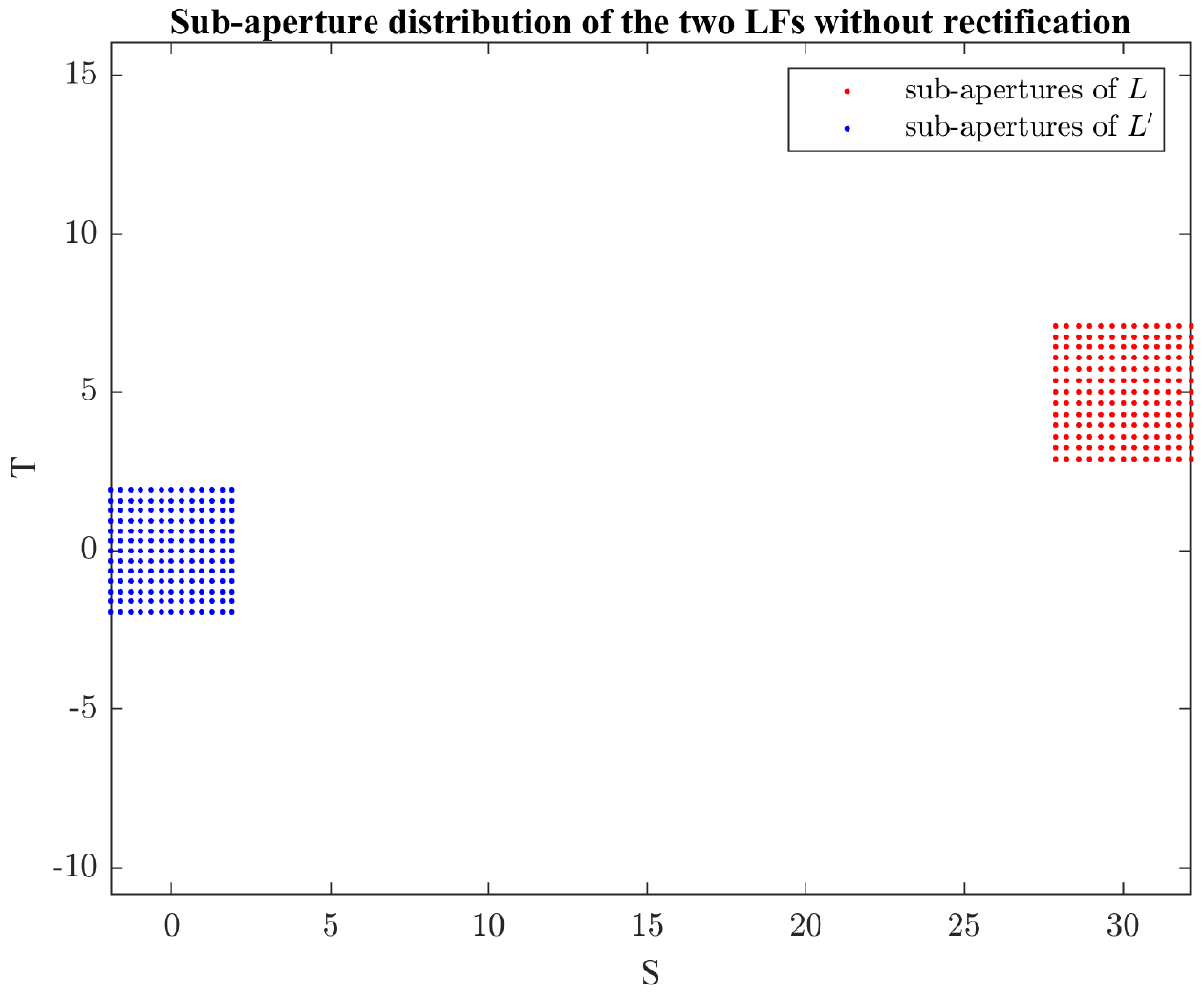}
	}
	\subcaptionbox{\label{fig:warping_b}}
	{
	\includegraphics[width=60mm]
	{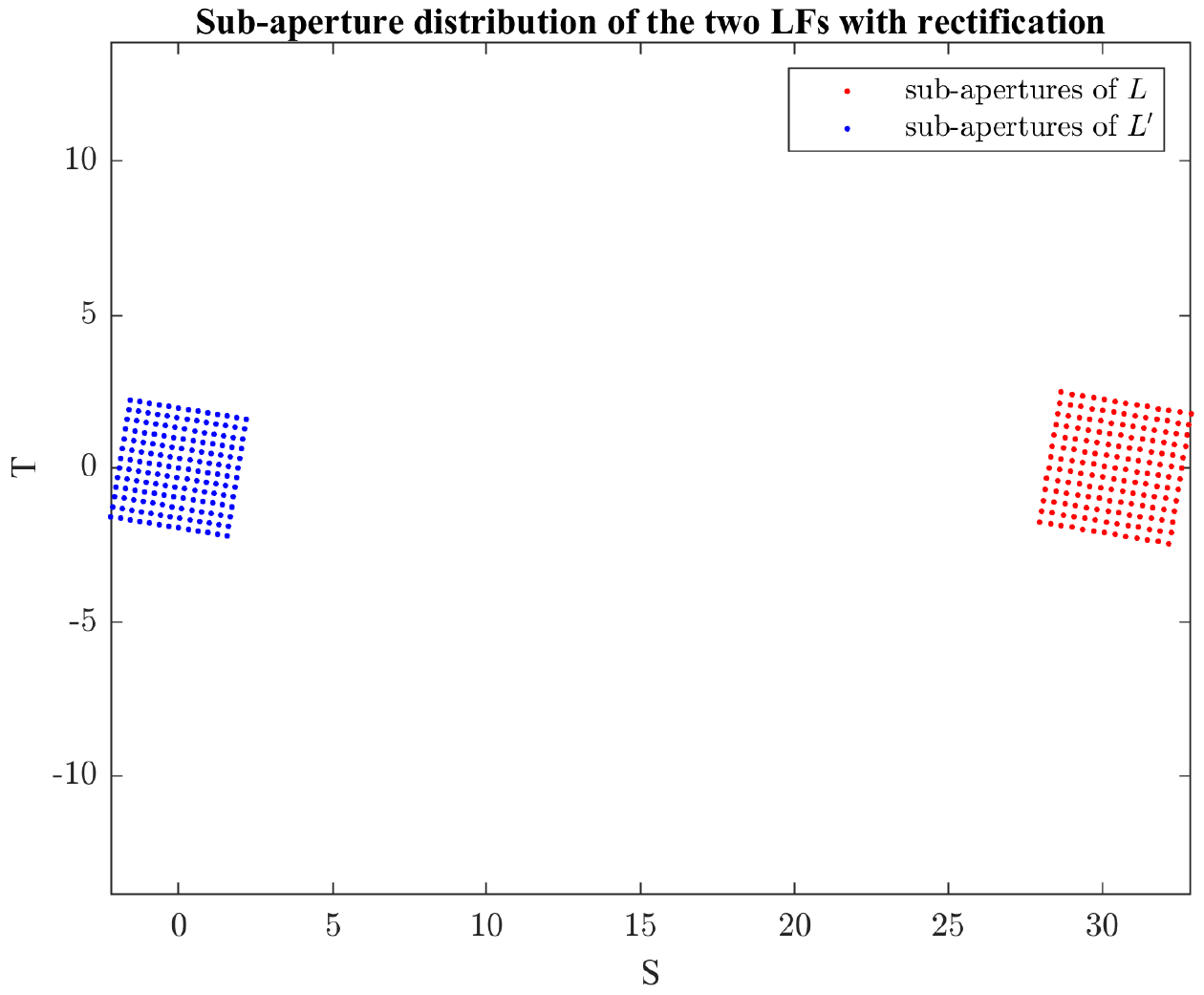}
	}
 \caption{The sub-aperture distribution of two LFs. (a)  without rectification; (b) with rectification.}
\label{fig:warping}
\end{figure}

Due to the randomness of the relative pose, the directly warped LF is still not effective in enlarging the baseline. To illustrate the underlying problem, we set two cameras that only have a 5mm vertical spacing and no rotation. Fig.~\ref{fig:warping} shows the warping result of two LFs with $13\times13$ sub-apertures each, where the red dots represent the sub-apertures of $L$, and the blue dots represent the sub-apertures of $L^\prime$. As shown in Fig.~\ref{fig:warping_a}, none of the sub-apertures from different LFs can be located on a horizontal line, which implies that the baseline was not enlarged. To solve this problem, we construct the third TPP and warp the rays of two LFs onto it.
 
 \subsection{Rectification matrix and vector}
Rectification first attempts to maximize the number of sub-apertures that can be located on horizontal lines. Specifically, we construct a rotation matrix $R_{rect}$ that can map the two center-view sub-apertures onto the third $ST$ plane and align them horizontally. To obtain the largest baseline, $R_{rect}$ is created by starting with the translation vector $T$. The first row $e_{1}^\top$ of the matrix is calculated as
 \begin{equation}
	e_{1}=\dfrac{T}{\|T\|}.
 \end{equation}
 
To make sure the center-view sub-aperture only have horizontal displacement, the next vector $e_{2}$ must be orthogonal to $e_{1}$. This can be accomplished by using the cross-product of $e_{1}$ with another direction. Selecting the vectorial sum of two principal rays can reduce reprojection distortion brought by warping two LFs. After normalization, another unit vector is calculated as
 \begin{equation}
	e_{2}=\operatorname{norm}\left(\left[\begin{matrix}
	-r_{2,3} t_{3}+t_{2}\left(r_{3,3}+1\right) \\
	r_{1,3} t_{3}-t_{1}\left(r_{3,3}+1\right) \\
	-r_{1,3} t_{2}+r_{2,3} t_{1}
	\end{matrix}\right]\right),
	\end{equation}
where $\operatorname{norm}(\cdot)$ represents the 2-norm of a vector.
In order to maintain the orthogonality of the rotation matrix, the third vector  $e_{3}$ orthogonal to $e_{1}$ and $e_2$ can be computed by using the cross-product operation as
 \begin{equation}
	e_{3}=e_{1} \times e_{2}.
\end{equation}
 
Then our matrix is calculated as
 \begin{equation}
	R_{r e c t}=\left[\begin{matrix}
	e_{1}^\top \\
	e_{2}^\top \\
	e_{3}^\top
	\end{matrix}\right].
	\end{equation}
 
The matrix rotates two $ST$ planes around their central sub-apertures, so that the two $ST$ planes can be coplanar, and the two central sub-apertures have the same vertical coordinate in one coordinate system. The warping of two LFs into the third TPP parameterization is implemented by setting $R_l=R_{rect}$ and $R_r=R_{rect} R$. $R_l$ and $R_r$ are the rotation matrices of the left camera and the right camera, respectively. Since there is no translation between the TPP of left camera and the third TPP, its translation vector  $T_{l}=\left[0,0,0\right]^\top$. The translation vector between the TPP of the right camera and the third TPP is calculated as $T_{r}=R_{rect} T$.
 
 \subsection{Interpolation of row-aligned SAIs}
\begin{figure}[htbp]
	\centering
	\includegraphics[width=90mm]
	{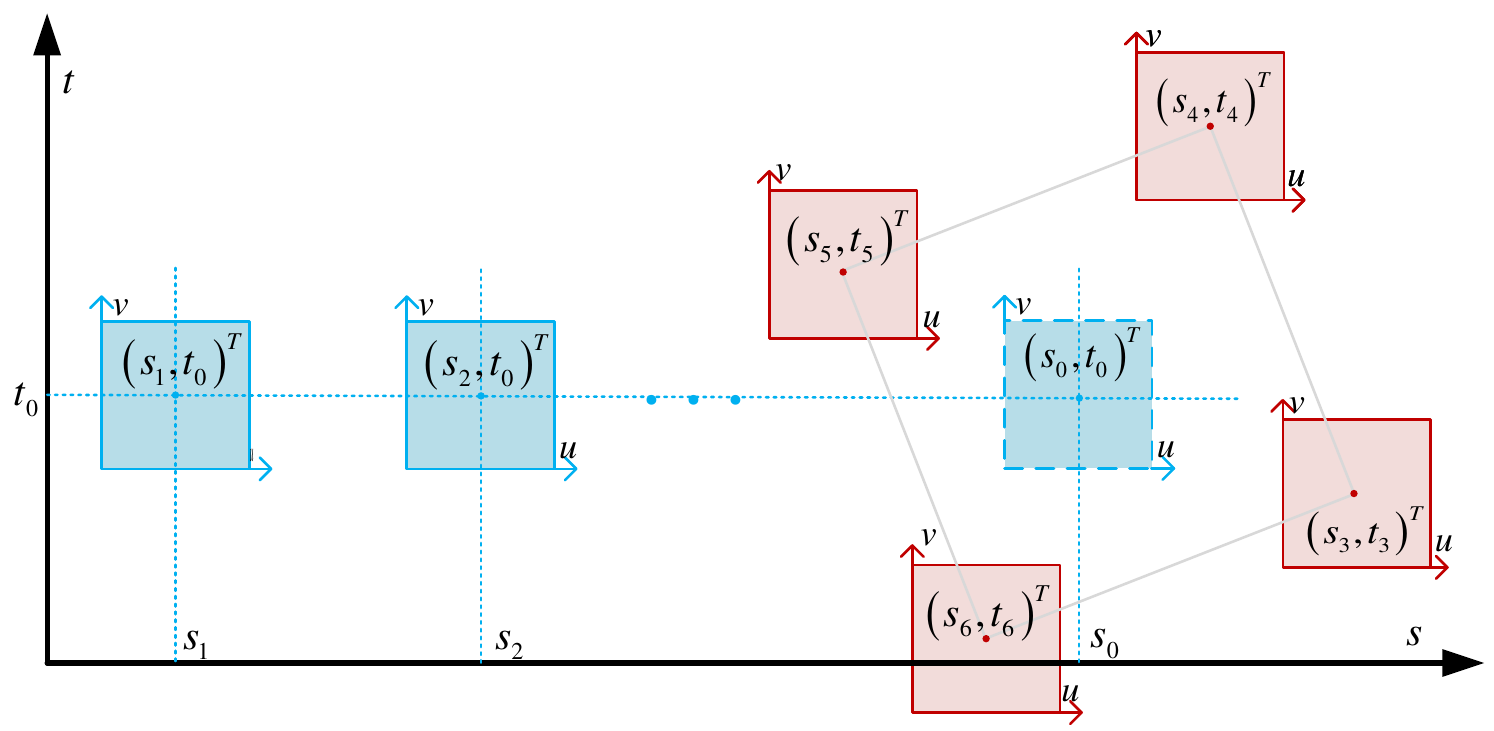}
	\caption{Interpolation on the regular grid.}
	\label{fig:interpo}
 \end{figure}

\begin{figure}[htbp]
	\centering
	\subcaptionbox{}{
	\includegraphics[width=95mm]
	{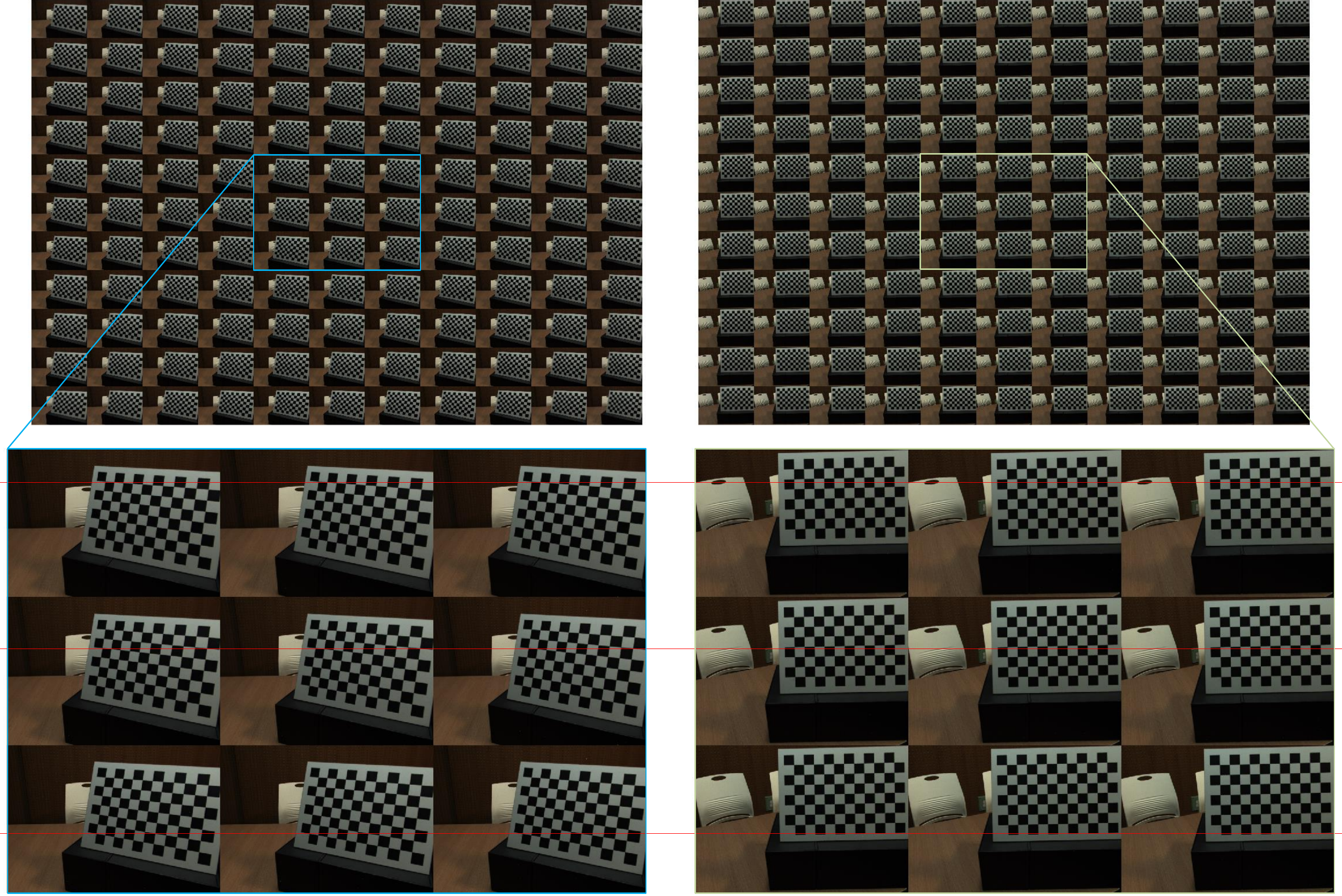}
	}
	\subcaptionbox{\label{fig:recedSAIsb}}{
	\includegraphics[width=102mm]
	{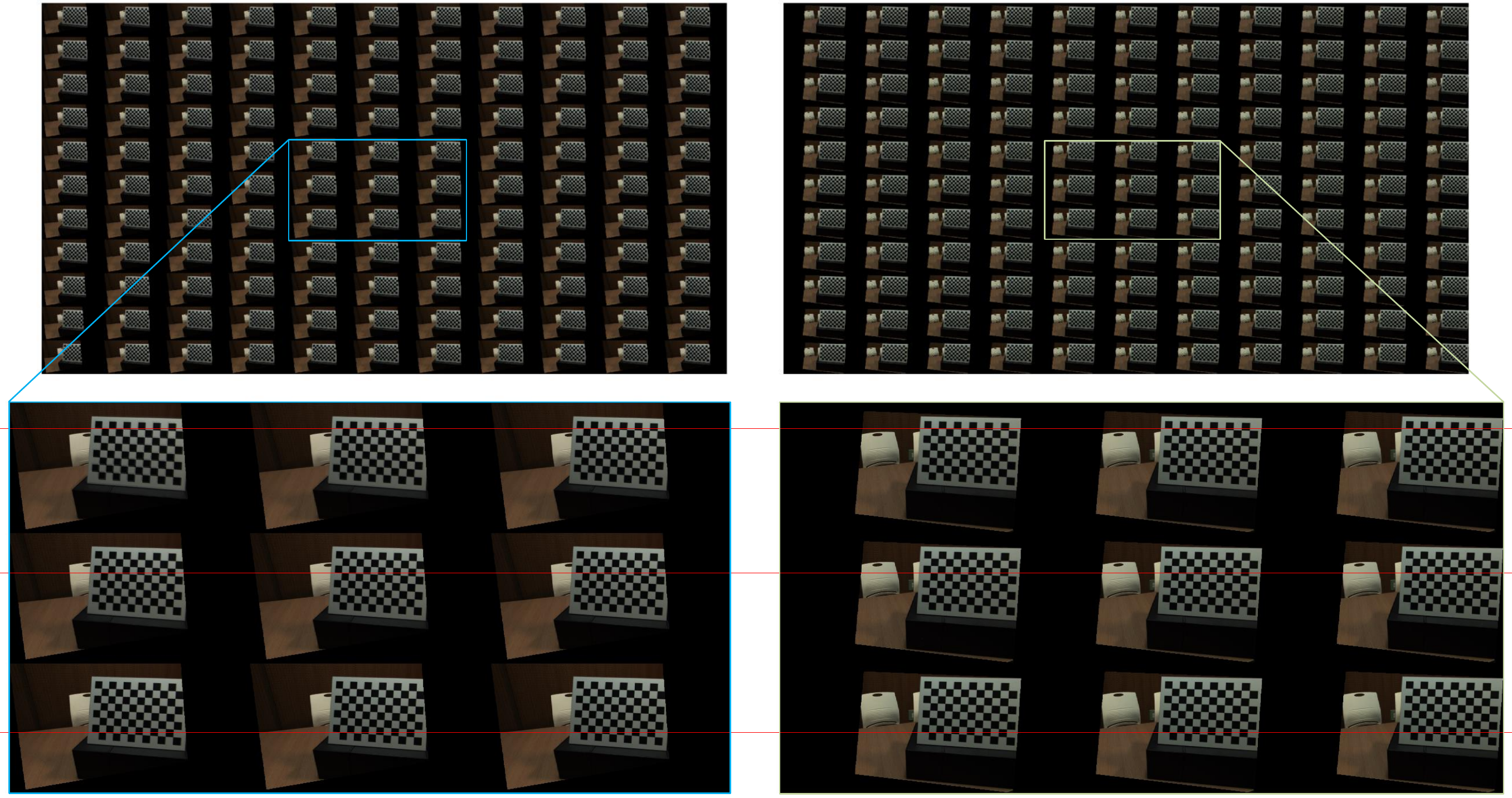}
	}
 \caption{Light field rectification: (a) original left and right SAIs pair (b) the rectified left and right SAIs pair; note that the scan lines are aligned in the rectified images}
 \label{fig:recedSAIs}
 \end{figure}

After transforming the rays of the two LFs onto the third TPP by the method described in Sec.~\ref{warping}, the distribution of sub-apertures in the third TPP are shown in Fig.~\ref{fig:warping_b}. Apparently, the number of sub-apertures from different LFs that can be located on the horizontal line increases significantly. In order to obtain the images of aligned sub-apertures, an interpolation as shown in Fig.~\ref{fig:interpo} is required. For each synthetic sub-aperture, such as $(s_0,t_0)$, which needs to be aligned, the luminance of each ray is acquired by calculating the ray coordinate in its original TPP and implementing the quadrilinear interpolation. This means that the surrounding apertures $(s_i,t_i)$ with $i=3,4,5,6$ are located, and in each of them, four rays are used to interpolate a ray. As shown in Fig.~\ref{fig:recedSAIs}, since the sub-apertures are aligned horizontally, corners in these SAIs are aligned in row. 
 
\begin{figure}[htbp]
	\centering
	\includegraphics[width=60mm]
	{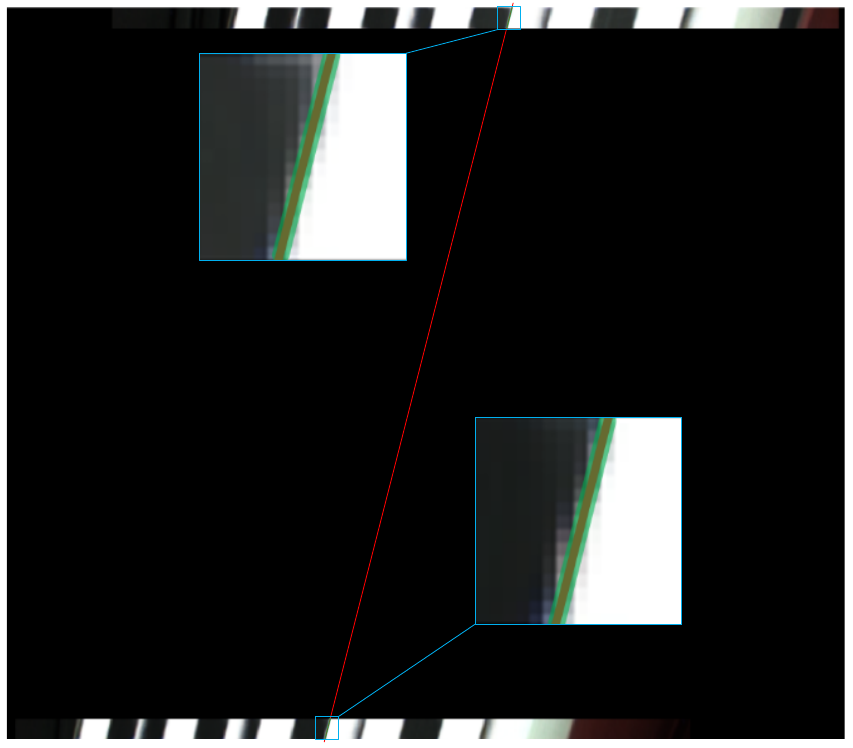}
	\caption{The EPI extracted from the rectified LF.}
	\label{fig:epi}
 \end{figure}

The corresponding EPI can also be extracted from the rectified LF as shown in Fig.~\ref{fig:epi}. A row of pixels in an EPI image corresponds to a row of pixels in a SAI. All pixel rows are extracted from SAIs in the same row in Fig.~\ref{fig:recedSAIsb}. Since the baseline is enlarged, the line on EPI becomes much longer. Although the two LFs span a long distance, the lines on EPI still match well, which is attributed to the accurate relative pose estimation.
 
\section{Experiment and analysis}
\subsection{Performance of pose estimation}
In order to evaluate the performance of the proposed algorithm for pose estimation, an MLA-based conventional LF camera is simulated, whose intrinsic parameters are listed in Table~\ref{tab1}. These parameters come from the setting of a Lytro Illum camera so as to obtain plausible input.

\begin{table}[htbp]
	\footnotesize
	\setlength{\abovecaptionskip}{0pt}
	\setlength{\belowcaptionskip}{0pt} 
	\centering
	\captionsetup{font={footnotesize,bf}}
	\caption{Intrinsic parameters of simulated plenoptic cameras (pixel)}
	\begin{tabular}{ccccccc}
  \hline
   & $f_x$ & $f_y$ & $c_x$ &$c_y$& $K_1$&$K_2$ \\
	  \hline
 camera 1 &572.720&572.685&	270.916&	188.109&	0.030	&165.298 \\
 camera 2 &538.374	&538.062	&283.471&188.709&	0.028 &147.606 \\
 
   \hline
   \end{tabular}
	\label{tab1}
\end{table}

For comparison purposes, we generate LF-points matches of checkerboard corners in two LFs based on different relative poses sets. LF-points of $7\times11$ corners with a spacing of 22.5mm under different checkerboard poses are generated. The geometry of generating is based on the correspondence between the LF-point and its 3D point. The correspondence model in~\cite{9694504} is used:
\begin{equation}\label{LFP-3DP}
\left[\begin{matrix}
u_{c} \\
v_{c} \\
\lambda \\
1
\end{matrix}\right]=\frac{1}{Z} \left[\begin{matrix}
f_{x} & 0 & c_{x} & 0 \\
0 & f_{y} & c_{y} & 0 \\
0 & 0 & -K_{1} & -K_{2} \\
0 & 0 & 1 & 0
\end{matrix}\right]\left[\begin{matrix}
X \\
Y \\
Z \\
1
\end{matrix}\right].
\end{equation}
According to the known corner coordinates $\left[X,Y,Z\right]^\top$ and the preset relative pose, the LF-points in the two LFs are generated respectively by Eq.~\eqref{LFP-3DP}. 

As claimed in Sec.~\ref{LFP}, all projection points of corners in $13\times13$ SAIs are calculated with corresponding LF-points. Independent Gaussian noise with mean zero and varying standard deviations $\sigma$ are added on the locations of these projected corners. For each noise level or pose configuration, 100 independent trials are performed and the average result is considered as the final performance. For fair comparison, suitable input for compared algorithms are computed using these projected points in SAI. The relative poses are set considering actual scenarios so that the corners can be observed in both LF cameras.

As the ground truth of relative pose is known, the performance of relative pose estimation algorithms can be evaluated by calculating the angular errors of relative poses which are represented as

\begin{equation}
   Err_{R_{a g l}}=\frac{180}{\pi} \times \operatorname{acos}\left(0.5 \times\left(\operatorname{trace}\left(R \times R_{e s t}^{\top}\right)-1\right)\right)
\end{equation}
and
\begin{equation}
   Err_{T_{a g l}}=\frac{180}{\pi} \times \operatorname{acos} \left(\frac{\operatorname{dot}\left(T, T_{e s t}\right)}{\operatorname{norm}(T) \times \operatorname{norm}\left(T_{e s t}\right)}\right),
\end{equation}
where $Err_{R_{a g l}}$ and $Err_{T_{a g l}}$ are angular errors of rotation matrix and translation vector, respectively. $\operatorname{acos}(\cdot)$ represents an arccosine function. $\operatorname{trace}(\cdot)$ calculates the trace of a matrix. $\operatorname{dot}(\cdot)$ represents the dot product operation. $R$ and $T$ are actual rotation matrix and translation vector. $R_{est}$ and $T_{est}$ are estimated rotation matrix and translation vector. $R_{est}^\top$ is the transpose of $R_{est}$. The performance of our proposed algorithm (Proposed) is compared with pose estimation algorithms~\cite{johannsen2015on} (LF-SfM),~\cite{9320357} (LF-DLT), and our previous work~\cite{9694504} (LFP). 

\subsubsection{Angular error w.r.t. the noise level}
In this experiment, the measurements of rotation angles set at $\left(5^\circ,20^\circ,5^\circ \right)$ and translation set at ${\left[{80,5,5} \right]^\top}$ are employed to verify the robustness of estimation algorithms. Standard deviation $\sigma$ varies from 0.1 to 0.5 pixel with 0.1 pixel step. As shown in Table~\ref{tab2}, higher noise level makes for larger errors on rotation matrix and translation vector for all algorithms. When noise is in low level, the errors of rotation and translation estimated by LF-SfM are relatively small. However, the errors grow rapidly and exceeds those estimated by other algorithms when noise increases gradually, which indicates the ray-based algorithm is not robust to noise. For LF-DLT, the error of translation estimated is almost always the largest and the error of rotation estimated increases relatively fast, since LF-DLT will be disturbed by the inaccurate 3D points reconstructed through the small baseline. In most cases, the proposed algorithm achieves the best performance and LFP ranks second to the proposed algorithm. In extreme cases  ($\sigma=2,3$), the error of proposed algorithm is significantly smaller than those estimated by the other algorithms, which indicates the robustness of the proposed algorithm.

  

\begin{table}[htbp]
	\footnotesize
	\setlength{\abovecaptionskip}{0pt}
	\setlength{\belowcaptionskip}{0pt} 
	\setlength\tabcolsep{3pt} 
	\centering
	\captionsetup{font={footnotesize,bf}}
	\caption{Angular Errors (degree) Under Different Noise Levels}
	\begin{tabular}{cccccccccc}
  \hline
   \multicolumn{2}{c}{}  & \multicolumn{4}{c}{Angular error of $R$}&\multicolumn{4}{c}{Angular error of $T$} \\
   \cmidrule(lr){3-6} \cmidrule(lr){7-10}
   \multicolumn{2}{c}{Algorithm}&LF-SfM & LF-DLT & LFP&Proposed&LF-SfM & LF-DLT &LFP&Proposed\\
  \hline
  
  \multirow{6}{*}{${\sigma}$}&0.1 & 0.0448&	0.0687&	0.0355&	\textbf{0.0275}&	\textbf{0.0649}	&	0.1687&	0.1587&	0.1355\\
  \multirow{6}{*}{(pixel)}&0.2&	0.1799&	0.2896&	0.0795&	\textbf{0.0789}&		0.3318&	0.3269 &	0.3312&\textbf{0.2997}\\
  &0.3 &0.3971&	0.4018&	0.1321&	\textbf{0.1264}&		0.7527&	1.0586 &	0.4965&	\textbf{0.4502}\\
  &0.4 &0.6758&	0.5266&	0.2134&	\textbf{0.1571}&	1.3145&	1.6824 &0.7697&\textbf{0.5578}\\
  &0.5 &	1.0112&	0.9291&	0.3417&	\textbf{0.2024}&	2.0125&	2.6552&	1.1309&	\textbf{0.7511}\\
  &2 &	4.1957&	3.7815&	2.3129&	\textbf{0.6415}&		10.4459&	11.0324&	8.2894&	\textbf{3.3611} \\
  &3 &	5.7847&	4.2603&	3.1094&	\textbf{0.9731}&		17.7719&	17.8734&	12.5036&	\textbf{4.5646} \\
   \hline
  \end{tabular}
  \label{tab2}
\end{table}

\subsubsection{Angular error w.r.t. different poses}
This experiment investigates the performance with respect to different relative poses. The rotation angles are $\left(5^\circ,15^\circ, 5^\circ\right)$($R_1$) and $\left(5^\circ, 30^\circ, 5^\circ\right)$($R_2$), and translation along x-axis are 50 mm ($T_1$) and 100mm ($T_2$), respectively.For each combination of rotation and translation, the independent Gaussian noise with zero mean and a standard deviation $\delta$ of 0.3 pixel is added. The results of performance are shown in Table~\ref{tab3}. When the rotation is fixed and translation increases, the angular errors of rotation and translation estimated with LF-SfM also increase, which indicates the ray-based algorithm is more suitable for small translation. The angular errors of translation estimated with LF-DLT are always the largest. Especially when the translation is small, the errors of LF-DLT are significantly larger than the other algorithms. The performance variation of LFP is the same as that of the proposed algorithm. When the rotation is fixed and translation increases, the errors of rotation increase significantly and the errors of translation change slightly. When the translation is fixed and rotation increases, the errors of rotation and translation also change slightly, which indicates the LF-feature-based algorithms are not sensitive to rotation and more suitable for small translation.
  

\begin{table}[htbp]
	\footnotesize
	\setlength{\abovecaptionskip}{0pt}
	\setlength{\belowcaptionskip}{0pt} 
	\setlength\tabcolsep{2.65pt} 
	\centering
	\captionsetup{font={footnotesize,bf}}
	\caption{Angular Errors (degree) Under Different Poses}
	\begin{tabular}{cccccccccc}
  \hline
   \multicolumn{2}{c}{}  & \multicolumn{4}{c}{Angular error of $R$}&\multicolumn{4}{c}{Angular error of $T$} \\
   \cmidrule(lr){3-6} \cmidrule(lr){7-10}
   \multicolumn{2}{c}{Algorithm}& LF-SfM & LF-DLT & LFP&Proposed& LF-SfM &LF-DLT &LFP&Proposed\\
  \hline
  
  \multirow{3}{*}{preset}&$R_1,T_1$ &0.2072&0.7678&	0.0958&	\textbf{0.0713}&		0.5533&	1.8685&	0.5924&\textbf{0.5083} \\
  \multirow{3}{*}{pose}&$R_1,T_2$&	0.4384&	0.4353&	0.1774&	\textbf{0.1340}&0.5763&	0.8721 &0.5113&	\textbf{0.4649}\\
  &$R_2,T_1$ &	0.1713&	0.8201&	0.0857&	\textbf{0.0628}&		0.7870&	1.7105 &	0.6185&	\textbf{0.5162}\\
  &$R_2,T_2$ &0.3741&0.5964&	0.1743&	\textbf{0.1237}&		0.8782&	0.9487 &	0.5765&	\textbf{0.4730}\\
   \hline
  \end{tabular}
  \label{tab3}
\end{table}

\subsection{Performance of rectification}
\begin{figure}[htbp]
	\centering
	\includegraphics[width=55mm]
	{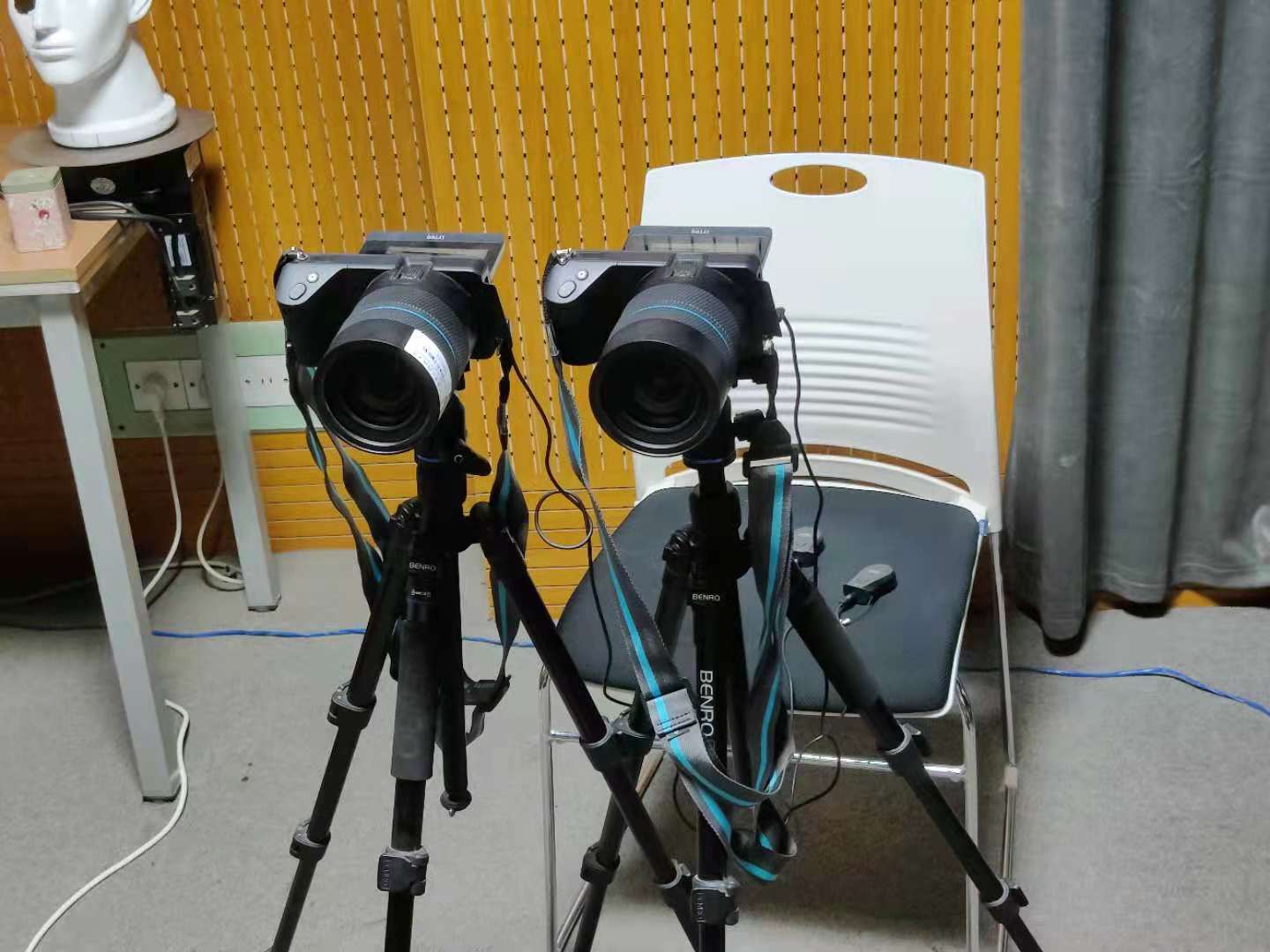}
	\caption{LF cameras location setup.}
	\label{fig:cameraSetup}
\end{figure}
\begin{figure}[htbp]
	\centering
	\subcaptionbox{}{
	\includegraphics[width=55mm]
	{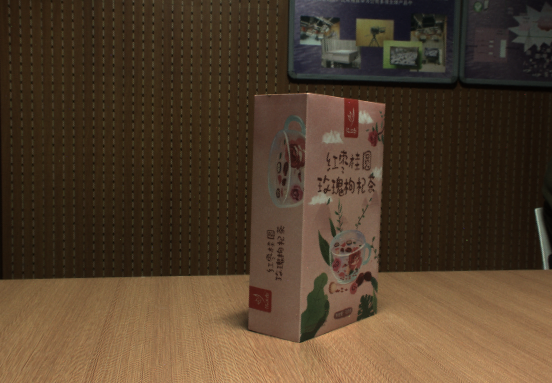}
	}
	\subcaptionbox{}{
	\includegraphics[width=55mm]
	{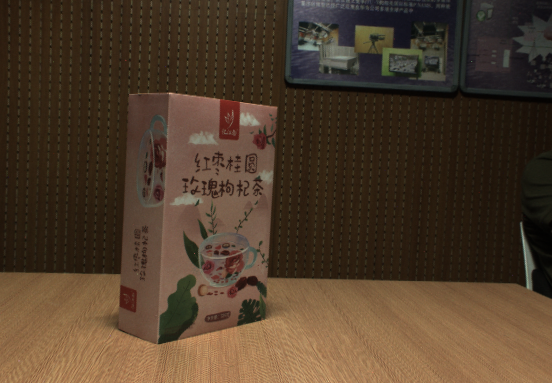}
	}
 \caption{Central SAIs. (a) central SAI of the left LF of Box and (b) central SAI of the right LF of Box}
 \label{CentralSAIs}
\end{figure}

In this section, the rectification performance and the scene reconstruction effects on real scenes are evaluated. We mount two calibrated Lytro Illum cameras in proper poses to make sure scenes are in the field of view of both cameras as shown in Fig.~\ref{fig:cameraSetup}. All real scenes are captured by the Lytro Illum cameras. Central SAIs of LFs are shown in Fig.~\ref{CentralSAIs}. Rectification of different pose estimation algorithms are implemented. After calibration, 3D point clouds can be recovered from the central SAI and its depth. For a single LF, we directly use its central SAI and depth estimated in~\cite{jeon2015accurate}. For the rectified LF, we choose one of the rectified central SAIs and use depth estimated in~\cite{jeon2015accurate}.

\subsubsection{Performance of rectification}
\begin{figure}[htbp]
	\centering
	\subcaptionbox{LF-SfM}{
	\includegraphics[height=45mm]
	{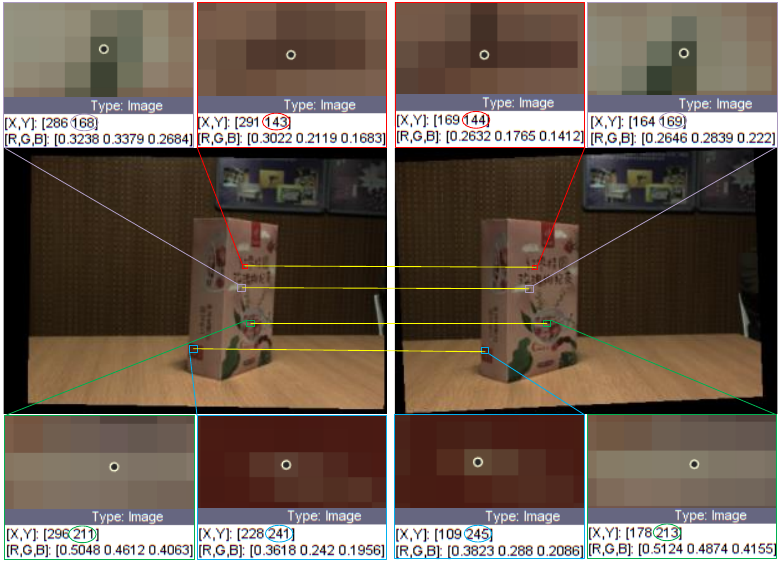}
	}
	\subcaptionbox{LF-DLT}{
	\includegraphics[height=45mm]
	{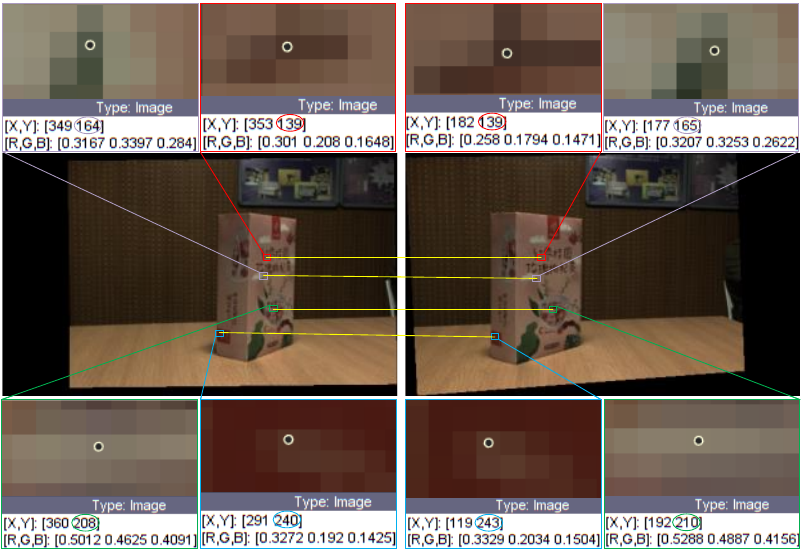}
	}
	\subcaptionbox{LFP}{
	\includegraphics[height=45mm]
	{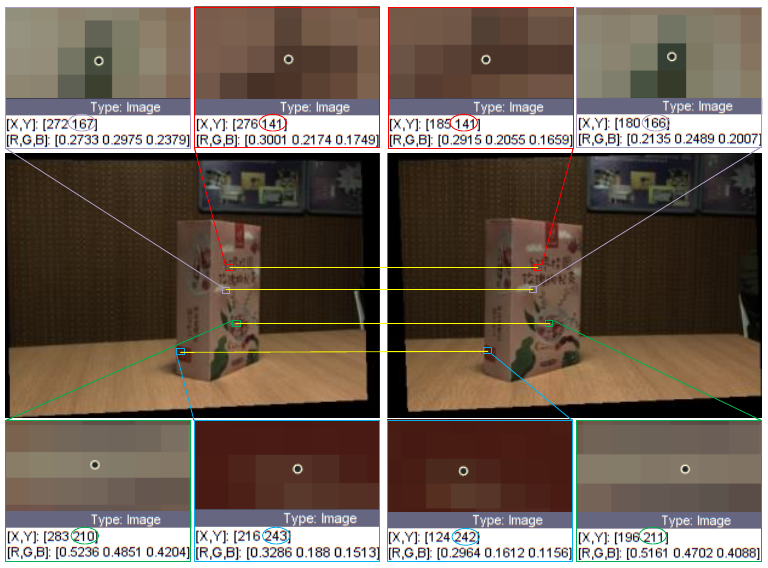}
	}
	\subcaptionbox{Proposed}{
	\includegraphics[height=45mm]
	{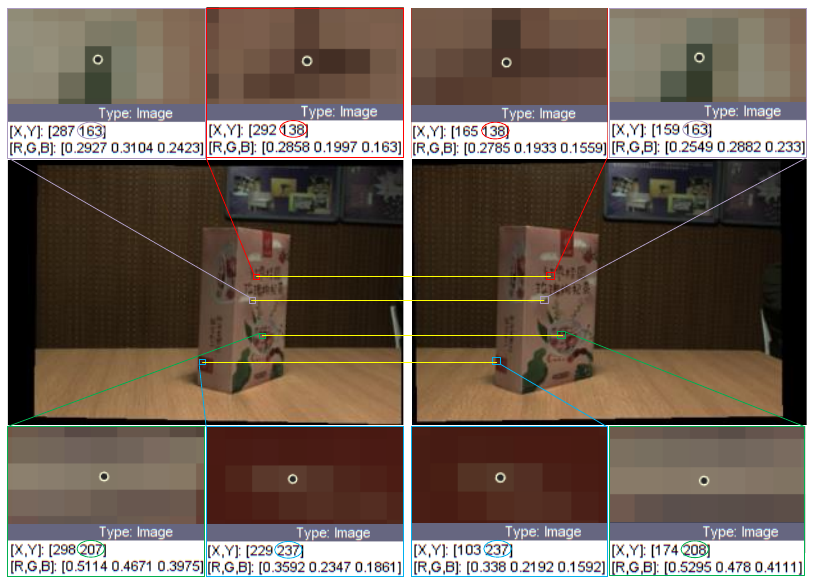}
	}
 \caption{Rectification performance.}
    \label{rec_exp}
\end{figure}

In this experiment, the rectified SAIs are shown to assess the performance of different algorithms on rectification. After rectification, the projected points of a scene point in the images of the same row of sub-apertures should have the same vertical coordinate. The rectified central SAI on each LF are shown in Fig.~\ref{rec_exp}. The vertical coordinates of feature points on each scene are marked on the images. From these results, we can see that the proposed pose estimation algorithm can achieve the minimum error on the vertical coordinate.

\subsubsection{Layered effect}
\begin{figure}[htbp]
	\centering
	\subcaptionbox{front view of the rectified LF}{
	\includegraphics[width=40mm]
	{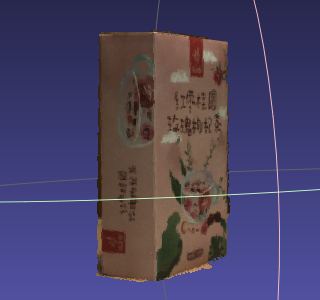}
	}
	\subcaptionbox{top view of the rectified LF}{
	\includegraphics[width=40mm]
	{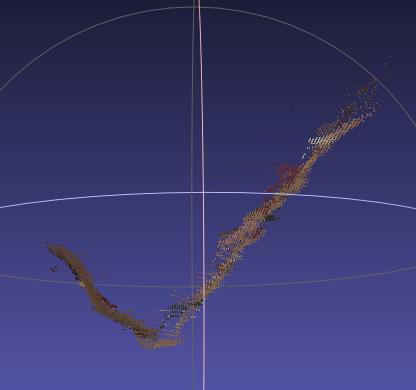}
	}
	\subcaptionbox{side view of the rectified LF}{
	\includegraphics[width=40mm]
	{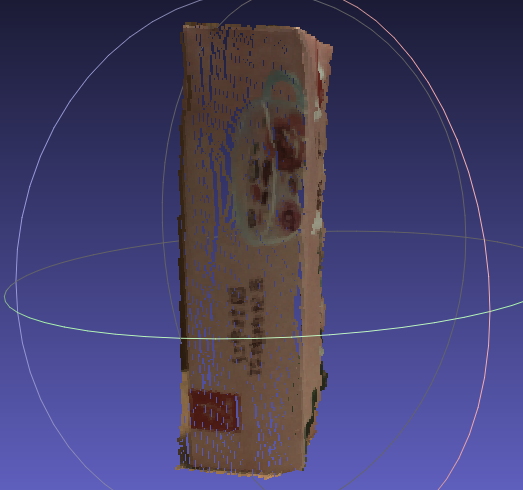}
	}
	\subcaptionbox{front view of the single LF}{
	\includegraphics[width=40mm]
	{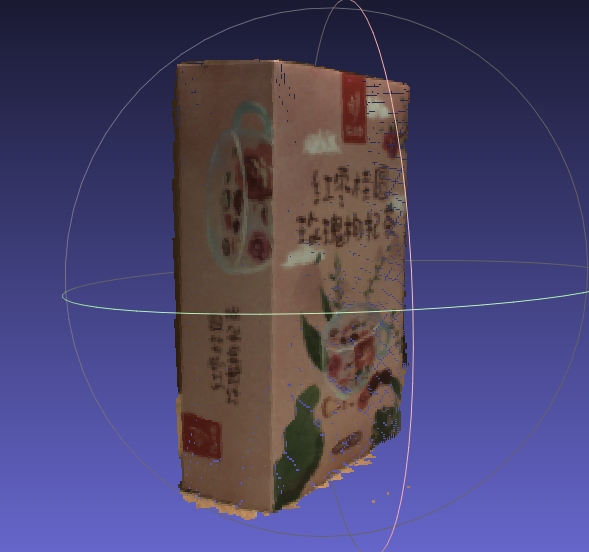}
	}
	\subcaptionbox{top view of the single LF}{
	\includegraphics[width=40mm]
	{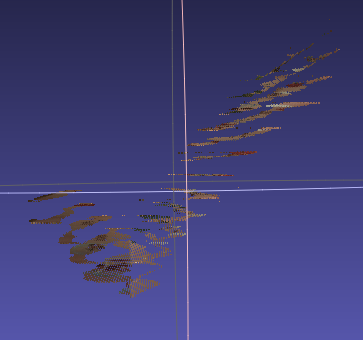}
	}
	\subcaptionbox{side view of the single LF}{
	\includegraphics[width=40mm]
	{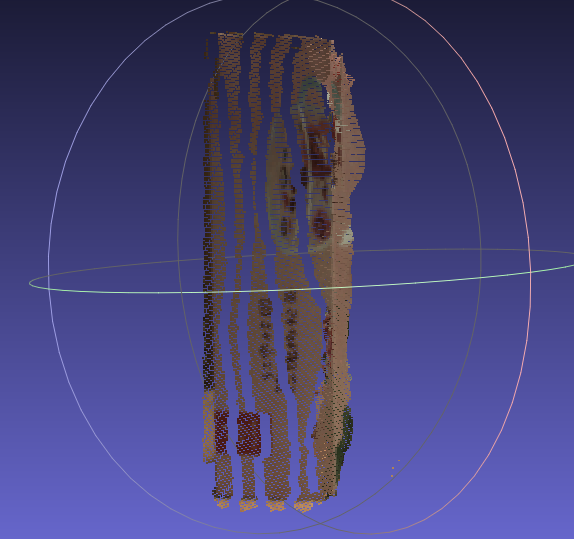}
	}
 \caption{Reconstruction from different views.}
 \label{layers}
\end{figure}

Due to the inherent characteristic of small baseline in hand-held LF camera, the disparity between adjacent SAIs is usually less than 1 pixel, which generally leads to layered effect in reconstructed point clouds. In stereo correspondence of LF depth estimation, first step is computing the matching cost, in which the disparity of each pixel is calculated through displacing the SAI and comparing the difference between it and other SAIs. In depth estimation in~\cite{jeon2015accurate}, the displacement can be sub-pixel, which are used to reduce the layered effect brought by the small baseline in substance. This experiment aims at comparing the layered effects on a single LF and the rectified LF to show the advantage of enlarging baseline. As shown in Fig.~\ref{layers}, point clouds with or without rectification both have a good outline of the object from the front view. However, when observed from side and top views, the layered effect of reconstructions in the single LF are very severe. On the other hand, the rectified LF shows no apparent layered effect.

\subsubsection{Distance measuring}
\begin{figure}[htbp]
	\centering
	\subcaptionbox{reconstruction of the box}{
	\includegraphics[height=50mm]
	{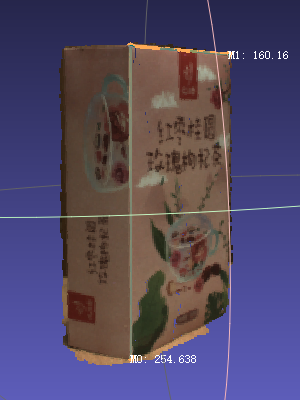}
	}
	\subcaptionbox{height measurement of the box}{
	\includegraphics[height=50mm]
	{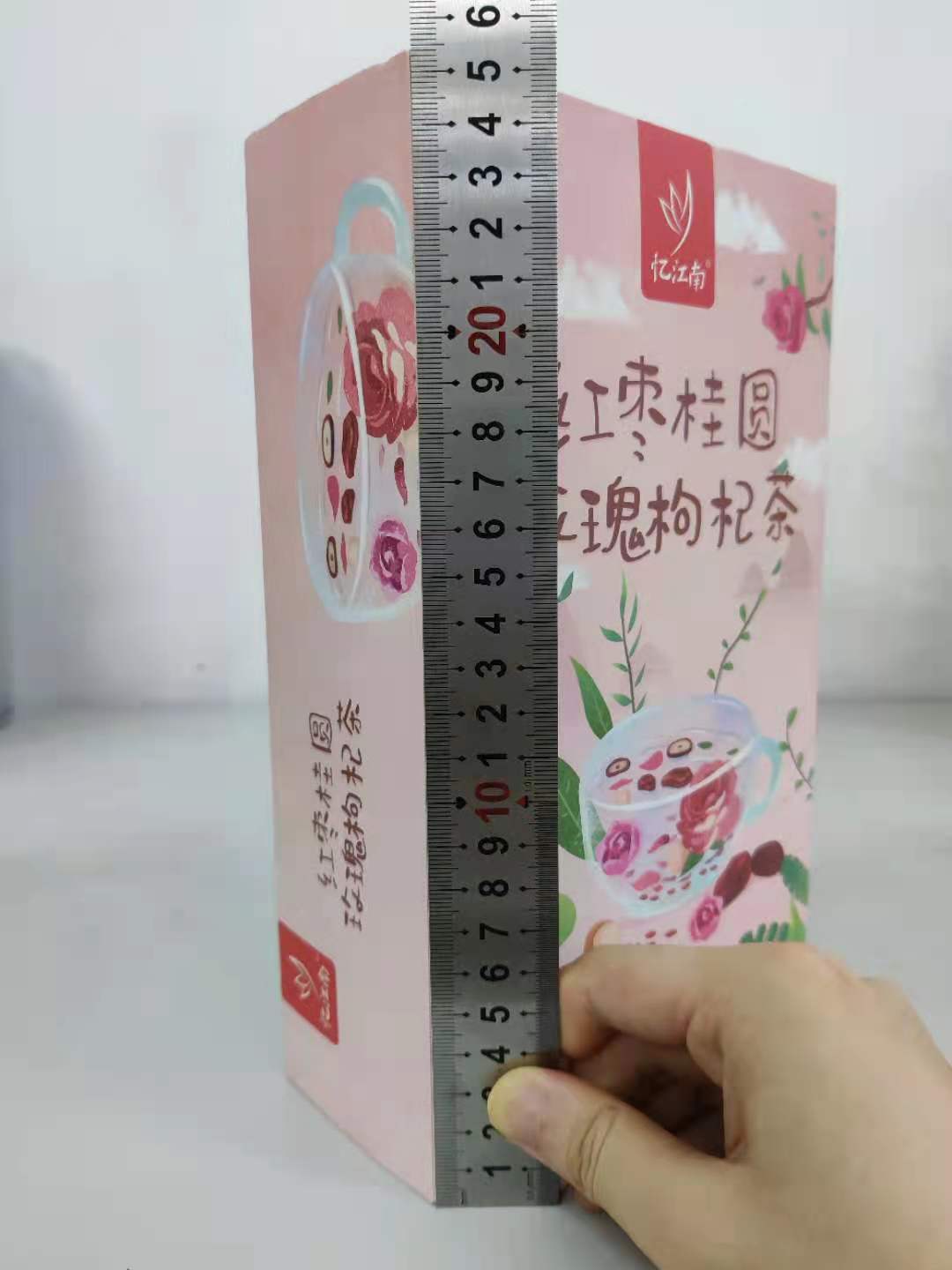}
	}
	\subcaptionbox{width measurement of the box}{
	\includegraphics[height=50mm]
	{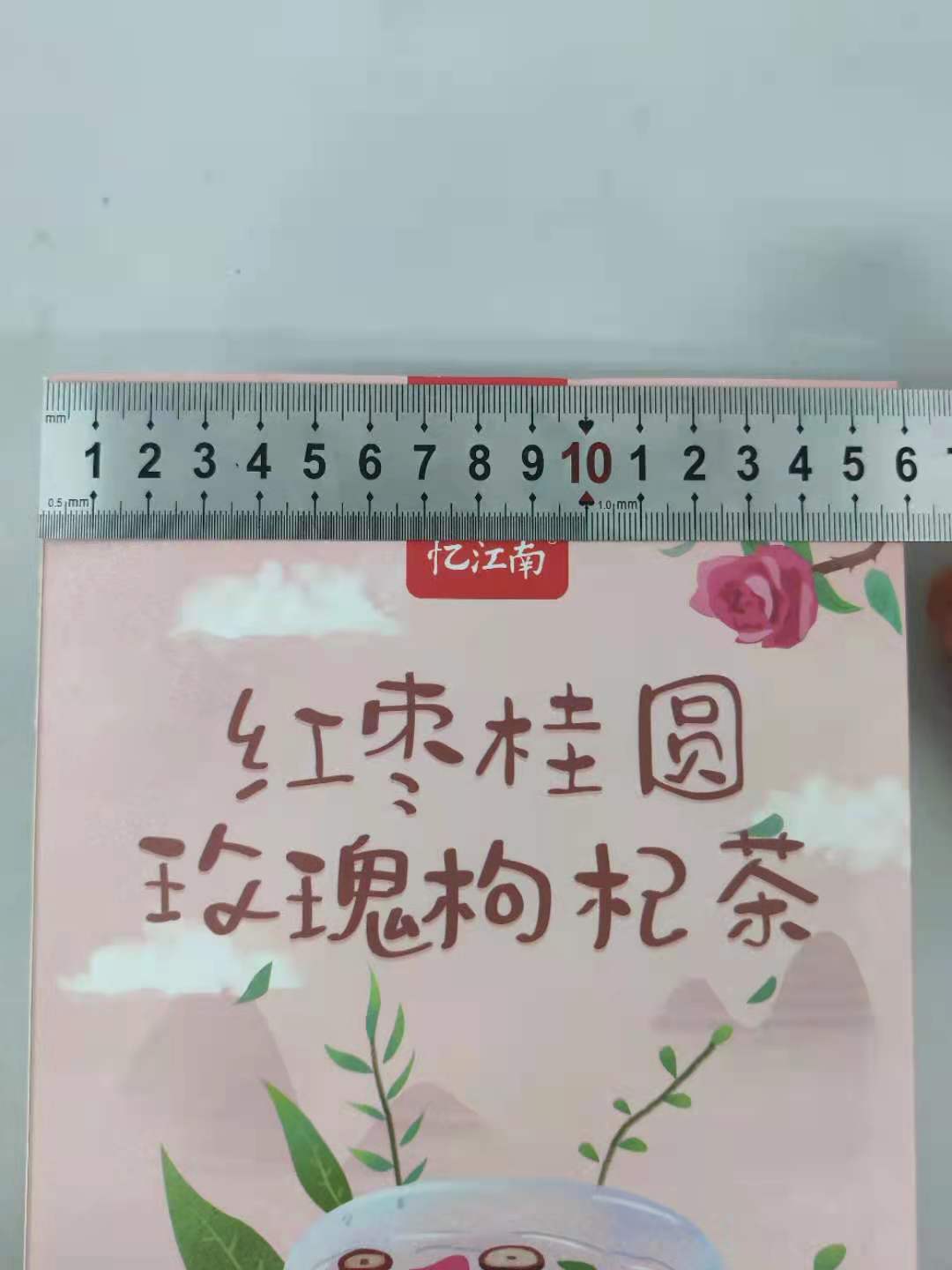}
	}
  \caption{Reconstruction and measurements of the box.}
  \label{recon}
\end{figure}

In this experiment, the metric reconstruction based on the rectified LF is implemented. The distances between several specific points are shown in Fig.~\ref{recon}, and the estimated distance between the reconstructed points are nearly equal to those measured lengths from real objects by a ruler. For these measurement examples, the accuracy of distance between reconstructed points demonstrates the effectiveness of the proposed method.

\section{conclusion}
In this paper, a method is proposed to rectify LFs captured by two hand-held LF cameras. In the relative pose estimation, we propose a linear approach to estimate the projective transformation between the different LF-point pairs and extract the relative pose. In rectification part, the constructed rectification matrix and vector are employed to align the two LFs, and through interpolation, the row-aligned SAIs in all sub-views can be obtained. These SAIs are applied in depth estimation to achieve depth with high depth resolution. The proposed pose estimation algorithm outperforms the classical and state-of-the-art algorithms, which shows the superiority on data format choosing and the accuracy on solution. The effectiveness of proposed method is also verified in real scenes.

\begin{backmatter}
\bmsection{Funding}
\bmsection{Acknowledgments}
The authors acknowledge supports from Fundamental Research Funds for the Central Universities (5001-20109215456) and  Innovation Fund of  Xidian Universitity (5001-20109215456).
\bmsection{Disclosures}
The authors declare no conflicts of interest.
\bmsection{Data Availability Statement}
Data underlying the results presented in this paper are not publicly available at this time but may be obtained from the authors upon reasonable request.
\bmsection{Supplemental document}
See Supplement 1 for supporting content. 

\end{backmatter}
\bibliography{sample}

\begin{thebibliography}{10}
\newcommand{\enquote}[1]{``#1''}

\bibitem{Lytroorg}
Lytro, \enquote{The lytro camera [online]. available,}
  \url{http://www.lytro.com/}. 2016.

\bibitem{Raytrixorg}
Raytrix, \enquote{3d light field camera technology [online]. available,}
  \url{http://www.raytrix.de/}. 2016.

\bibitem{dong2013plenoptic}
F.~Dong, S.-H. Ieng, X.~Savatier, R.~Etienne-Cummings, and R.~Benosman,
  \enquote{Plenoptic cameras in real-time robotics,} {\protect\JournalTitle{The
  International Journal of Robotics Research}} \textbf{32}, 206--217 (2013).

\bibitem{zeller2017calibration}
N.~Zeller, F.~Quint, and U.~Stilla, \enquote{From the calibration of a
  light-field camera to direct plenoptic odometry,} {\protect\JournalTitle{IEEE
  Journal of selected topics in signal processing}} \textbf{11}, 1004--1019
  (2017).

\bibitem{johannsen2016layered}
O.~Johannsen, A.~Sulc, N.~Marniok, and B.~Goldluecke, \enquote{Layered scene
  reconstruction from multiple light field camera views,} in \emph{Asian
  Conference on Computer Vision,}  (Springer, 2016), pp. 3--18.

\bibitem{Wu_2017_CVPR}
G.~Wu, M.~Zhao, L.~Wang, Q.~Dai, T.~Chai, and Y.~Liu, \enquote{Light field
  reconstruction using deep convolutional network on epi,} in \emph{Proceedings
  of the IEEE Conference on Computer Vision and Pattern Recognition (CVPR),}
  (2017).

\bibitem{Nousias_2019_CVPR}
S.~Nousias, M.~Lourakis, and C.~Bergeles, \enquote{Large-scale, metric
  structure from motion for unordered light fields,} in \emph{Proceedings of
  the IEEE/CVF Conference on Computer Vision and Pattern Recognition (CVPR),}
  (2019).

\bibitem{ng2005fourier}
R.~Ng, \enquote{Fourier slice photography,} in \emph{ACM transactions on
  graphics (TOG),}  vol.~24 (ACM, 2005), pp. 735--744.

\bibitem{10.1145/2980179.2980251}
N.~K. Kalantari, T.-C. Wang, and R.~Ramamoorthi, \enquote{Learning-based view
  synthesis for light field cameras,} {\protect\JournalTitle{ACM Trans.
  Graph.}} \textbf{35} (2016).

\bibitem{Jin_2020_CVPR}
J.~Jin, J.~Hou, J.~Chen, and S.~Kwong, \enquote{Light field spatial
  super-resolution via deep combinatorial geometry embedding and structural
  consistency regularization,} in \emph{Proceedings of the IEEE/CVF Conference
  on Computer Vision and Pattern Recognition (CVPR),}  (2020).

\bibitem{meng2020high}
N.~Meng, X.~Wu, J.~Liu, and E.~Lam, \enquote{High-order residual network for
  light field super-resolution,} in \emph{Proceedings of the AAAI Conference on
  Artificial Intelligence,}  vol.~34 (2020), pp. 11757--11764.

\bibitem{jeon2015accurate}
H.-G. Jeon, J.~Park, G.~Choe, J.~Park, Y.~Bok, Y.~Tai, and I.~So~Kweon,
  \enquote{Accurate depth map estimation from a lenslet light field camera,} in
  \emph{Proceedings of the IEEE conference on computer vision and pattern
  recognition,}  (2015), pp. 1547--1555.

\bibitem{tao2013depth}
M.~W. Tao, S.~Hadap, J.~Malik, and R.~Ramamoorthi, \enquote{Depth from
  combining defocus and correspondence using light-field cameras,} in
  \emph{Proceedings of the IEEE International Conference on Computer Vision,}
  (2013), pp. 673--680.

\bibitem{wanner2012globally}
S.~Wanner and B.~Goldluecke, \enquote{Globally consistent depth labeling of 4d
  light fields,} in \emph{2012 IEEE Conference on Computer Vision and Pattern
  Recognition,}  (IEEE, 2012), pp. 41--48.

\bibitem{8985549}
K.~Mishiba, \enquote{Fast depth estimation for light field cameras,}
  {\protect\JournalTitle{IEEE Transactions on Image Processing}} \textbf{29},
  4232--4242 (2020).

\bibitem{Huang_2017_ICCV}
C.-T. Huang, \enquote{Robust pseudo random fields for light-field stereo
  matching,} in \emph{Proceedings of the IEEE International Conference on
  Computer Vision (ICCV),}  (2017).

\bibitem{rogge2020depth}
S.~Rogge, I.~Schiopu, and A.~Munteanu, \enquote{Depth estimation for
  light-field images using stereo matching and convolutional neural networks,}
  {\protect\JournalTitle{Sensors}} \textbf{20}, 6188 (2020).

\bibitem{8263242}
H.-G. Jeon, J.~Park, G.~Choe, J.~Park, Y.~Bok, Y.-W. Tai, and I.~S. Kweon,
  \enquote{Depth from a light field image with learning-based matching costs,}
  {\protect\JournalTitle{IEEE Transactions on Pattern Analysis and Machine
  Intelligence}} \textbf{41}, 297--310 (2019).

\bibitem{bolles1987epipolarp}
R.~C. Bolles, H.~H. Baker, and D.~H. Marimont, \enquote{Epipolar-plane image
  analysis: An approach to determining structure from motion,}
  {\protect\JournalTitle{International Journal of Computer Vision}} \textbf{1},
  7--55 (1987).

\bibitem{wilburn2005high}
B.~Wilburn, N.~Joshi, V.~Vaish, E.-V. Talvala, E.~Antunez, A.~Barth, A.~Adams,
  M.~Horowitz, and M.~Levoy, \enquote{High performance imaging using large
  camera arrays,} in \emph{ACM Transactions on Graphics (TOG),}  vol.~24 (ACM,
  2005), pp. 765--776.

\bibitem{1315176}
B.~Wilburn, N.~Joshi, V.~Vaish, M.~Levoy, and M.~Horowitz, \enquote{High-speed
  videography using a dense camera array,} in \emph{Proceedings of the 2004
  IEEE Computer Society Conference on Computer Vision and Pattern Recognition,
  2004. CVPR 2004.},  vol.~2 (2004), pp. II--II.

\bibitem{1211520}
R.~Pless, \enquote{Using many cameras as one,} in \emph{2003 IEEE Computer
  Society Conference on Computer Vision and Pattern Recognition, 2003.
  Proceedings.},  vol.~2 (2003), pp. II--587.

\bibitem{4587545}
H.~Li, R.~Hartley, and J.~hak Kim, \enquote{A linear approach to motion
  estimation using generalized camera models,} in \emph{2008 IEEE Conference on
  Computer Vision and Pattern Recognition,}  (2008), pp. 1--8.

\bibitem{johannsen2015on}
O.~Johannsen, A.~Sulc, and B.~Goldluecke, \enquote{On linear structure from
  motion for light field cameras,} in \emph{international conference on
  computer vision(ICCV),}  (2015), pp. 720--728.

\bibitem{9320357}
S.~Nousias, M.~Lourakis, P.~Keane, S.~Ourselin, and C.~Bergeles, \enquote{A
  linear approach to absolute pose estimation for light fields,} in \emph{2020
  International Conference on 3D Vision (3DV),}  (2020), pp. 672--681.

\bibitem{9694504}
S.~Zhang, D.~Jin, Y.~Dai, and F.~Yang, \enquote{Relative pose estimation for
  light field cameras based on lf-point-lf-point correspondence model,}
  {\protect\JournalTitle{IEEE Transactions on Image Processing}} pp. 1--1
  (2022).

\bibitem{2020arXiv200103734J}
D.~{Jin}, S.~{Zhang}, X.~{Huo}, W.~{Zhang}, and F.~{Yang}, \enquote{{A Two-step
  Calibration Method for Unfocused Light Field Camera Based on Projection Model
  Analysis},} {\protect\JournalTitle{arXiv e-prints}} arXiv:2001.03734v2
  (2021).

\bibitem{bok2016geometric}
Y.~Bok, H.-G. Jeon, and I.~Kweon, \enquote{Geometric calibration of
  micro-lens-based light field cameras using line features,}
  {\protect\JournalTitle{IEEE transactions on pattern analysis and machine
  intelligence}} \textbf{39}, 287--300 (2016).

\bibitem{levoy1996light}
M.~Levoy and P.~Hanrahan, \enquote{Light field rendering,} in \emph{Proceedings
  of the 23rd annual conference on Computer graphics and interactive
  techniques,}  (1996), pp. 31--42.

\bibitem{gortler1996lumigraph}
S.~J. Gortler, R.~Grzeszczuk, R.~Szeliski, and M.~F. Cohen, \enquote{The
  lumigraph,} in \emph{Proceedings of the 23rd annual conference on Computer
  graphics and interactive techniques,}  (1996), pp. 43--54.

\bibitem{hartley2003multiple}
R.~Hartley and A.~Zisserman, \emph{Multiple view geometry in computer vision}
  (Cambridge university press, 2003).

\bibitem{Horn:87}
B.~K.~P. Horn, \enquote{Closed-form solution of absolute orientation using unit
  quaternions,} {\protect\JournalTitle{J. Opt. Soc. Am. A}} \textbf{4},
  629--642 (1987).

\bibitem{strasdat2010real}
H.~Strasdat, J.~Montiel, and A.~J. Davison, \enquote{Real-time monocular slam:
  Why filter?} in \emph{2010 IEEE International Conference on Robotics and
  Automation,}  (IEEE, 2010), pp. 2657--2664.

\bibitem{engel2014lsd}
J.~Engel, T.~Sch{\"o}ps, and D.~Cremers, \enquote{Lsd-slam: Large-scale direct
  monocular slam,} in \emph{European conference on computer vision,}
  (Springer, 2014), pp. 834--849.

\bibitem{zeller2018scale}
N.~Zeller, F.~Quint, and U.~Stilla, \enquote{Scale-awareness of light field
  camera based visual odometry,} in \emph{Proceedings of the European
  Conference on Computer Vision (ECCV),}  (2018), pp. 715--730.

\end{thebibliography}
\end{document}